\documentclass[final]{cvpr}

\usepackage{times}
\usepackage{epsfig}
\usepackage{graphicx}
\usepackage{amsmath}
\usepackage{amssymb}
\usepackage{caption,subcaption}
\usepackage{pifont}
\usepackage{colortbl}
\usepackage{booktabs}
\usepackage[table]{xcolor}
\usepackage[etex=true,export]{adjustbox}

\newtheorem{theorem}{Remark}
\definecolor{mygreen}{RGB}{0,127,128}   %007F80
\definecolor{mylight}{RGB}{167,44,167}  %A72CA7
\definecolor{pearDark}{RGB}{41, 128, 185}
\usepackage[pagebackref=true,breaklinks=true,colorlinks,bookmarks=false]{hyperref}

\newlength\savewidth\newcommand\shline{\noalign{\global\savewidth\arrayrulewidth
  \global\arrayrulewidth 1pt}\hline\noalign{\global\arrayrulewidth\savewidth}}

\def\cvprPaperID{1108} % *** Enter the CVPR Paper ID here
\def\confYear{CVPR}

\definecolor{Gray}{gray}{0.9}
\definecolor{blizzardblue}{rgb}{0.67, 0.9, 0.93}

\begin{document}

%%%%%%%%% TITLE
\title{Region Rebalance for Long-Tailed Semantic Segmentation}
\author{
	Jiequan Cui $^{1}$ \quad 
	Yuhui Yuan $^{2}$ \quad
	Zhisheng Zhong $^{1}$ \quad
	Zhuotao Tian $^{1}$ \quad
	Han Hu $^{2}$ \quad
	Stephen Lin $^{2}$ \quad
	Jiaya Jia $^{1}$ \\
	$^{1}$The Chinese University of Hong Kong \hspace{1cm} $^{2}$Microsoft Research Asia \hspace{1cm} 
    \vspace{.7em}\\
	{\tt\small \{jqcui, zszhong21, zttian, leojia\}@cse.cuhk.edu.hk, \{yuhui.yuan, hanhu, stevelin\}@microsoft.com}
}

\maketitle

%%%%%%%%% ABSTRACT
\begin{abstract}
In this paper, we study the problem of class imbalance in semantic segmentation.
We first investigate and identify the main challenges of addressing this issue through pixel rebalance. Then a simple and yet effective region rebalance scheme is derived based on our analysis. In our solution, pixel features belonging to the same class are grouped into region features, and a rebalanced region classifier is applied via an auxiliary region rebalance branch during training.
To verify the flexibility and effectiveness of our method,
we apply the region rebalance module into various semantic segmentation methods, such as Deeplabv$3$+, OCRNet, and Swin.
Our strategy achieves consistent improvement on the challenging ADE$20$K and COCO-Stuff benchmark.
In particular, with the proposed region rebalance scheme, state-of-the-art \textsc{BEiT} receives +\textbf{$0.7\%$} gain in terms of $\mathrm{mIoU}$ on the ADE$20$K \texttt{val} set.
\end{abstract}

%%%%%%%%% BODY TEXT
\section{Introduction}
Deep neural networks (DNNs) like convolutional neural networks \cite{alexnet,he2016deep,vggnet,googlenet} and vision transformers \cite{dosovitskiy2020image, touvron2021training, liu2021swin} have achieved great success in various computer vision tasks, including image classification \cite{alexnet,he2016deep,vggnet,googlenet}, object detection \cite{ren2015faster,DBLP:conf/cvpr/LinDGHHB17,DBLP:conf/cvpr/LiuQQSJ18} and semantic segmentation \cite{DBLP:conf/cvpr/ZhaoSQWJ17, chen2018encoder, xiao2018unified, yuan2020object}.
Most previous studies focus on curated data with annotations, such as ImageNet, that are balanced over different classes. In contrast, real-world data often exhibit a long-tailed distribution where a small number of head classes contain many instances while most other classes have relatively few instances.
As shown in~\cite{Gupta2019LVIS}, addressing the long-tailed distribution problem is the key to large vocabulary vision recognition tasks.

\begin{figure}[t]
\begin{center}
\hspace{-7mm}
\includegraphics[width=0.49\textwidth]{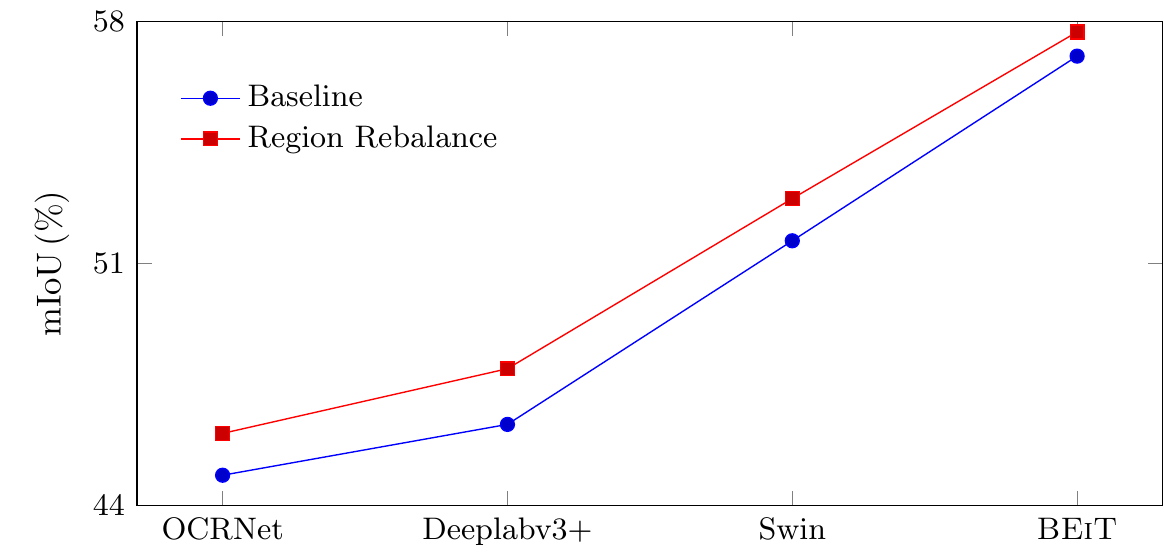}
\caption{\textbf{Illustrating the improvements with our region rebalance:}
incorporating our region rebalance scheme leads to +$1.21\%$, +$1.61\%$, +$1.22\%$, and +$0.7\%$ gains for OCRNet, DeepLabv$3$+, Swin, and \textsc{BEiT} on the ADE$20$K \texttt{val} set, respectively.}
\label{fig:comparison_ade20k}
\end{center}
\vspace{-0.3in}
\end{figure}

With the increasing attention on the long-tailed distribution problem, various advanced methods have been developed for long-tailed image classification \cite{buda2018systematic,huang2016learning,DBLP:conf/cvpr/CuiJLSB19, he2009learning,chawla2002smote,DBLP:conf/nips/RenYSMZYL20,DBLP:conf/iclr/KangXRYGFK20, DBLP:journals/corr/abs-2010-01809, DBLP:journals/corr/abs-2101-10633, DBLP:conf/nips/TangHZ20, Zhong_2021_CVPR,cui2021parametric} and instance segmentation \cite{tan2020eql, li2020overcoming, wang2021seesaw}.
For example, Cao et al. \cite{ldam} and Ren et al. \cite{DBLP:conf/nips/RenYSMZYL20} provide theoretical foundation on the optimal margins for long-tailed image classification. For long-tailed instance segmentation, most effort tackles the problem based on the Mask-RCNN framework \cite{he2017mask} and conducts class rebalance in the proposal classification.
Without much prior long-tailed study for semantic segmentation yet, we focus our work in this direction.

To address the long-tailed semantic segmentation problem, we first apply the well-known long-tailed image classification method to the semantic segmentation task. With the balanced softmax strategy \cite{DBLP:conf/nips/RenYSMZYL20} to rebalance the pixel
classification, we observe that it does not work well for the major evaluation metric of $\mathrm{IoU}$.

From an investigation of this, we identify two major challenges in conducting pixel rebalance:
(i) Pixel rebalance mainly improves pixel accuracy but not $\mathrm{IoU}$.
We empirically observe that applying the pixel rebalance strategy even harms $\mathrm{IoU}$ performance and argue that inconsistency between the training objective, {\textit i.e.}, pixel cross-entropy, and our inference target, {\textit i.e.}, $\mathrm{IoU}$, leads to this phenomenon.
(ii) Neighboring pixels are highly correlated. In long-tailed image classification, rebalance can be effectively guided by class frequency because image samples are independent and identically distributed (i.i.d.). In contrast, pixel samples are not i.i.d. due to correlation among neighboring pixels in an image. As a result, class frequency in the pixel domain is an unsuitable guide for rebalancing.

\begin{figure*}[t]
	\begin{center}
		\includegraphics[width=1.0\textwidth]{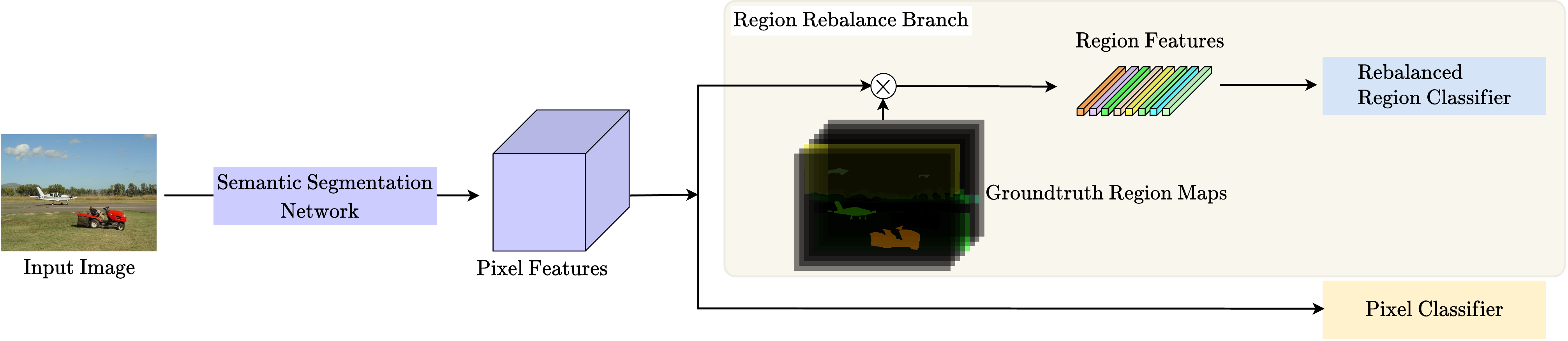}
		\caption{\textbf{Illustrating the region rebalance scheme for semantic segmentation:} The region rebalance branch is indicated by the tan-colored box. We introduce the region rebalance branch (w/o any other changes to the semantic segmentation network and pixel classifier) to handle the class imbalance problem during training and remove it during evaluation.}
		\label{fig:RRT_method}
	\end{center}
	\vspace{-0.00in}
\end{figure*}

In this work, we propose to overcome the non-i.i.d. issue of pixel rebalance by gathering correlated pixels into {\em regions} and performing rebalance based on regions. We refer to this method for dealing with the class imbalance in semantic segmentation as \emph{Region Rebalance}. In this approach, we introduce an auxiliary region classification branch where pixel features of the same class/region are averaged and fed into a rebalanced region classifier, as shown in Figure~\ref{fig:RRT_method}. With this additional training target, the pixel features are encouraged to lie in a more balanced classification space with regard to uncorrelated pixel samples. We note that the region rebalance branch is used only for training and is removed for inference.

An incidental benefit of this approach is that popular benchmark datasets, including ADE20K and COCO-Stuff164K, exhibit less class imbalance in the region domain than in the pixel domain, as later analyzed in Sec.~\ref{sec:datasets_stats}. Because of this property, rebalancing in the region domain can be accomplished relatively more effectively.

We apply our region rebalance strategy to
various semantic segmentation methods {\it e.g.}, PSPNet~\cite{DBLP:conf/cvpr/ZhaoSQWJ17}, OCRNet~\cite{yuan2020object}, DeepLabv$3$+~\cite{chen2018encoder}, Swin~\cite{liu2021swin}, and \textsc{BEiT}~\cite{bao2021beit}, and conduct experiments on two challenging semantic segmentation benchmarks, namely ADE$20$K\cite{zhou2017scene} and COCO-Stuff$164$K\cite{caesar2018coco},
to verify the effectiveness of our method.
The results are reported in
Figure \ref{fig:comparison_ade20k}.
Our code will be made publicly available.
The key contributions are summarized as follows:
\begin{itemize}
    \vspace{-0.1in}
	\item We investigate the class imbalance problem that exists in semantic segmentation and identify the main challenges of pixel rebalance.
	\vspace{-0.1in}
	\item We propose a simple yet effective region rebalance scheme and validate the effectiveness of our method across various semantic segmentation methods. Notably, our method yields a +$0.7\%$ improvement for \textsc{BEiT} \cite{bao2021beit}, setting a new performance record on ADE$20$K \texttt{val} at the time of submission. 
\end{itemize}

\section{Related Work}
\subsection{Long-tailed Image Classification}
In the area of long-tailed image classification, the most popular methods for dealing with imbalanced data can be categorized into re-weighting/re-sampling methods \cite{shen2016relay, zhong2016towards, buda2018systematic, byrd2019effect, he2009learning, japkowicz2002class, DBLP:journals/nn/BudaMM18, DBLP:conf/cvpr/HuangLLT16,huang2019deep,wang2017learning, ren2018learning, shu2019meta, jamal2020rethinking} or one-stage/two-stage methods \cite{cao2019learning, kang2019decoupling, Zhong_2021_CVPR, zhou2019bbn,cui2021parametric,DBLP:journals/corr/abs-2101-10633}.

\vspace{1mm}
\noindent\textbf{Re-weighting/Re-sampling}.
Re-sampling approaches are based on either over-sampling low-frequency classes or under-sampling high-frequency classes. Over-sampling \cite{shen2016relay, zhong2016towards, buda2018systematic, byrd2019effect} usually suffers from heavy over-fitting to low-frequency classes. For under-sampling \cite{he2009learning, japkowicz2002class, DBLP:journals/nn/BudaMM18}, it inevitably leads to degradation of CNN generalization ability because a large portion of the high-frequency class data is discarded.
Re-weighting \cite{DBLP:conf/cvpr/HuangLLT16,huang2019deep,wang2017learning, ren2018learning, shu2019meta, jamal2020rethinking} loss functions is an alternative to rebalancing and is accomplished by enlarging weights
on more challenging or sparse classes. However, re-weighting makes CNNs difficult to optimize when training on large-scale data.

\vspace{1mm}
\noindent\textbf{One-stage/Two-stage}.
Since the deferred re-weighting and re-sampling strategies proposed by Cao et al. \cite{cao2019learning}, it was further observed by Kang et al. \cite{kang2019decoupling} and Zhou et al. \cite{zhou2019bbn} that re-weighting and re-sampling could benefit classifier learning but hurt representation learning. Based on this, many two-stage methods \cite{cao2019learning, kang2019decoupling, Zhong_2021_CVPR} were developed.

The two-stage design contradicts the push towards end-to-end learning that has been prevalent in the deep learning era. Through a causal inference framework, Tang et al. \cite{tang2020long} revealed the harmful causal effects of SGD momentum on long-tailed classification. Ren et al. \cite{DBLP:conf/nips/RenYSMZYL20} extended LDAM \cite{cao2019learning} and theoretically obtained the optimal margins for multi-class image classification. Cui et al. \cite{DBLP:conf/cvpr/CuiJLSB19} proposed a residual learning mechanism to address this issue. Recently, contrastive learning has also been introduced for long-tailed image classification \cite{cui2021parametric}.

\subsection{Long-tailed Instance Segmentation}
With the great progress in long-tailed image classification \cite{buda2018systematic,huang2016learning,DBLP:conf/cvpr/CuiJLSB19, he2009learning,chawla2002smote,DBLP:conf/nips/RenYSMZYL20,DBLP:conf/iclr/KangXRYGFK20, DBLP:journals/corr/abs-2010-01809, DBLP:journals/corr/abs-2101-10633, DBLP:conf/nips/TangHZ20, Zhong_2021_CVPR,cui2021parametric}, many researchers have started to explore the long-tailed phenomenon in instance segmentation. Many algorithms \cite{tan2020eql, li2020overcoming, wang2021seesaw} have been developed to address this issue.

Tan et al. \cite{tan2020eql} observed that each positive sample of one category could be seen as a negative sample for other categories, causing the tail categories to receive more discouraging gradients. Based on this, they proposed to simply ignore those gradients for rare categories. 
Li et al. \cite{li2020overcoming} proposed balanced group softmax (BAGS) for balancing the classifier within a detection framework through group-wise training. BAGS implicitly modulates the training process for the head and tail classes and ensures that all classes are trained sufficiently. Wang et al. \cite{wang2021seesaw} proposed the Seesaw loss to dynamically rebalance gradients of positive and negative samples for each category, further improving performance.

\subsection{Semantic Segmentation}
Semantic segmentation has long been studied as a fundamental task in computer vision. Substantial progress was made with the introduction of Fully Convolutional Networks (FCN) \cite{long2015fully}, which formulate the semantic segmentation task as per-pixel classification. Subsequently, many advanced methods have been introduced.

Many approaches \cite{amirul2017gated,badrinarayanan2017segnet,lin2017refinenet, pohlen2017full, ronneberger2015u} combine upsampled high-level feature maps and low-level feature maps to capture global information and recover sharp object boundaries. A large receptive field also plays an important role in semantic segmentation. Several methods \cite{chen2017deeplab,chen2014semantic,yu2015multi} have proposed dilated or atrous convolutions to enlarge the field of filters and incorporate larger context. For better global information, recent work \cite{chen2017rethinking, chen2018encoder, DBLP:conf/cvpr/ZhaoSQWJ17} adopted spatial pyramid pooling to capture multi-scale contextual information. Along with atrous spatial pyramid pooling and an effective decoder module, Deeplabv$3$+ \cite{chen2018encoder} features a simple and effective encoder-decoder architecture.

\section{Our Method}
First, we analyze the challenges of conducting pixel rebalance for semantic segmentation.
Second, we present the details of our proposed region rebalance scheme and explain how region rebalance helps.

\subsection{Challenges of Pixel Rebalance}
\label{sec:challenges_rebalance}

\subsubsection{Higher Pixel Accuracy $\to$ Higher $\mathrm{IoU}$?} 
\label{sec:challenge1}
In the image classification task, minimizing the cross-entropy loss essentially maximizes the probability of the ground truth label during the training phase, which is consistent with the top-$1$ accuracy evaluation metric during inference.
Therefore, we can expect more balanced class accuracy via incorporating rebalance strategies.

For semantic segmentation, we typically use pixel cross-entropy as a proxy to indirectly optimize Intersection-over-Union ($\mathrm{IoU}$). However, pixel cross-entropy inherently maximizes pixel accuracy ($\mathrm{Acc}$). This misalignment between the training objective and target evaluation metric $\mathrm{IoU}$ inspires us to explore what would happen when rebalancing with pixel cross-entropy in semantic segmentation.

\begin{figure*} 
	\begin{center}
	    \begin{minipage}[t]{0.32\textwidth} 
			\begin{center}
				\includegraphics[width=0.99\textwidth]{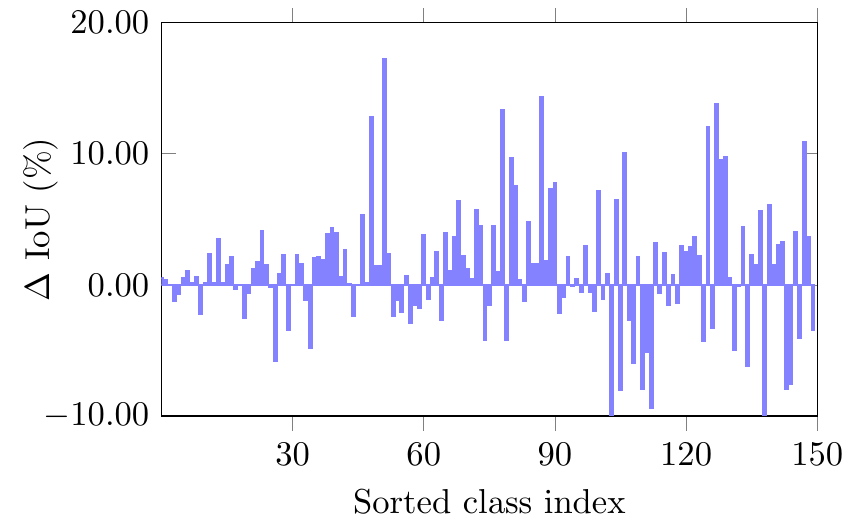} 
				\subcaption{$\Delta \mathrm{IoU}$ after rebalancing with our Region Rebalance method.} 
			\end{center} 
		\end{minipage} 
		\hspace{0.05in}
		\begin{minipage}[t]{0.32\textwidth} 
			\begin{center}
				\includegraphics[width=0.99\textwidth]{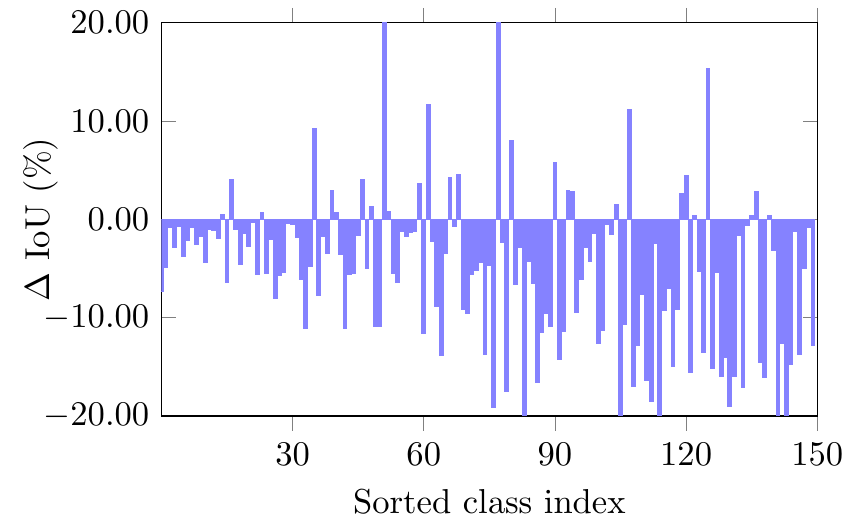} 
				\subcaption{$\Delta \mathrm{IoU}$ after pixel rebalance with balanced softmax \cite{DBLP:conf/nips/RenYSMZYL20} }
			\end{center} 
		\end{minipage} 
		\hspace{0.05in}
		\begin{minipage}[t]{0.32\textwidth} 
			\begin{center}
				\includegraphics[width=0.99\textwidth]{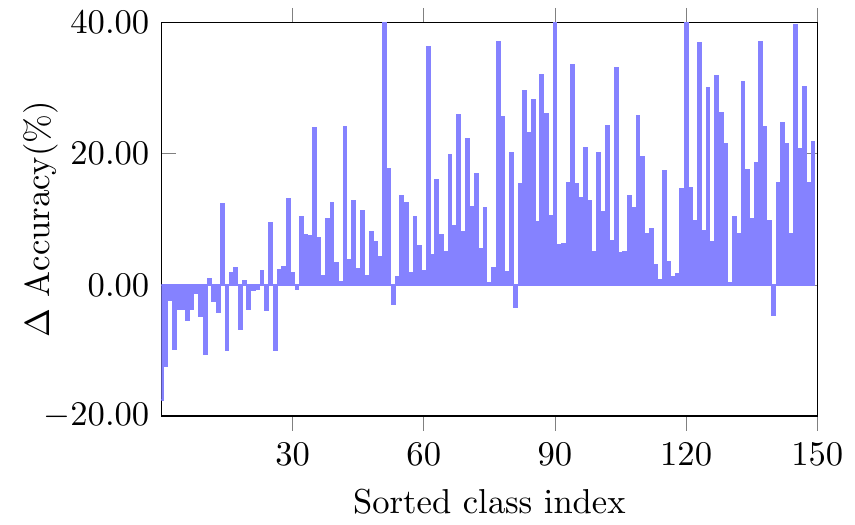} 
				\subcaption{$\Delta$Acc after pixel rebalance with balanced softmax \cite{DBLP:conf/nips/RenYSMZYL20}} 
			\end{center}
		\end{minipage}
	\end{center}
	\vspace{-0.20in}
	\caption{Comparison between baseline and pixel rebalance with balanced softmax. Classes are sorted in descending order of pixel numbers.}
	\label{fig:baseline_vs_pixrebalance}
\end{figure*}

\begin{figure*}[t]
	\begin{center}
		\begin{minipage}[t]{0.32\textwidth} 
			\begin{center}
				\includegraphics[width=0.99\textwidth]{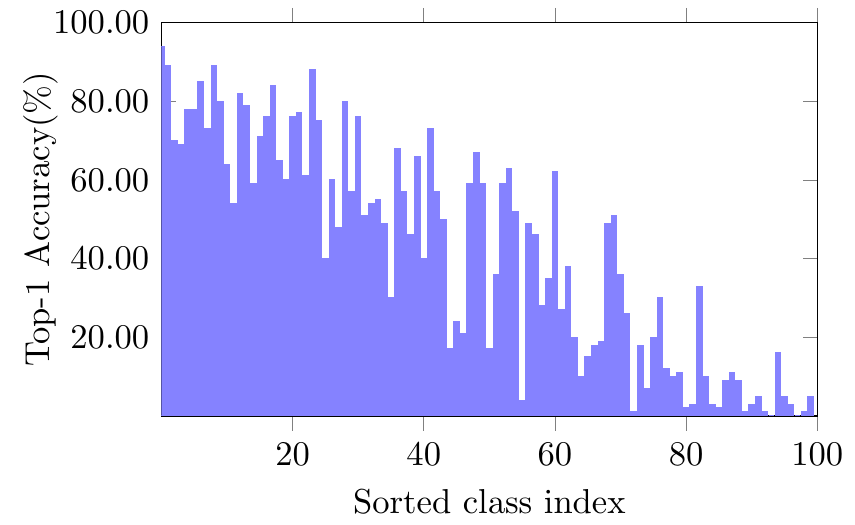} 
				\subcaption{Accuracy on CIFAR-$100$-LT with imbalance ratio 100.} 
				\label{fig:cifar100_acc} 
			\end{center} 
		\end{minipage} 
		\hspace{0.05in}
		\begin{minipage}[t]{0.32\textwidth} 
			\begin{center}
				\includegraphics[width=0.99\textwidth]{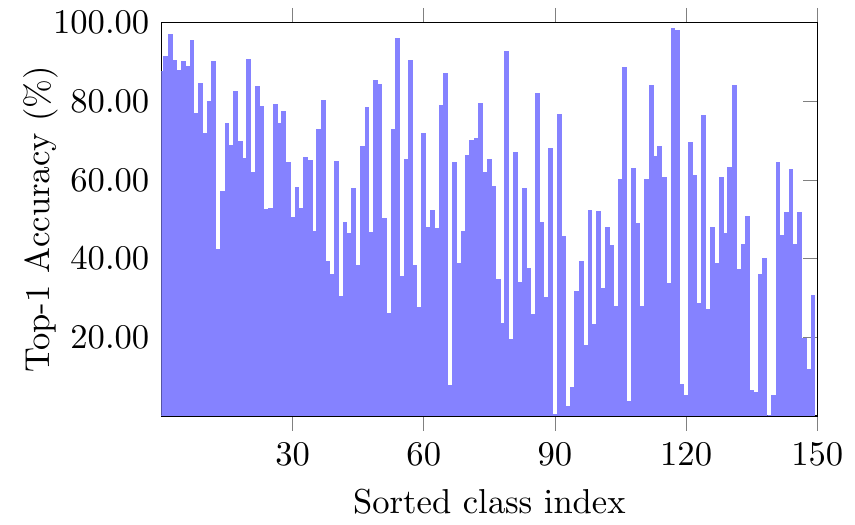} 
				\subcaption{Pixel accuracy on ADE$20$K.} 
				\label{fig:ade20k_pixel_acc} 
			\end{center}
		\end{minipage}
		\hspace{0.05in}
		\begin{minipage}[t]{0.32\textwidth} 
			\begin{center}
				\includegraphics[width=0.99\textwidth]{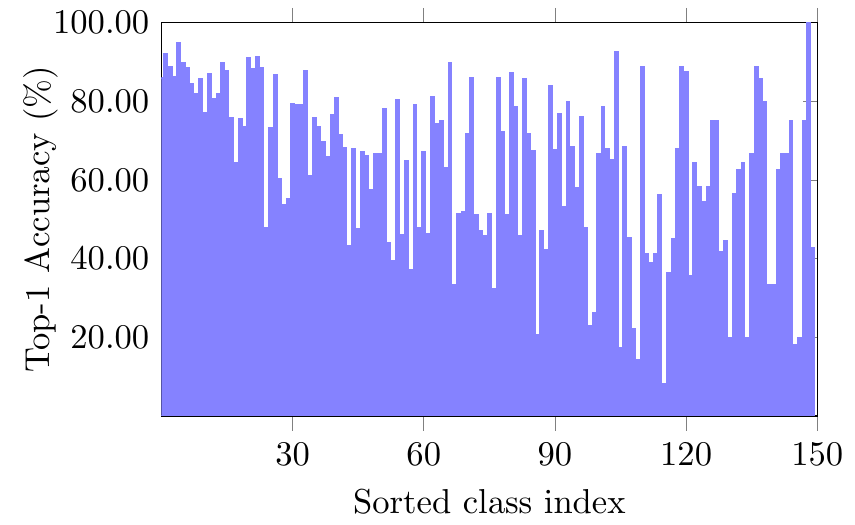} 
				\subcaption{Region accuracy on ADE$20$K.} 
				\label{fig:ade20k_img_acc} 
			\end{center}
		\end{minipage}
	\end{center}
	\vspace{-0.20in}
	\caption{Class accuracy is positively correlated with class image frequency in image classification, while class accuracy has weak correlations with class pixel frequency due to correlations among neighboring pixels. Class indexes are sorted in descending order according to the number of images, pixels, or regions in the classes.}
	\label{fig:accs_cifar100_ade20k}
\end{figure*}

\vspace{1mm}
\noindent\textbf{Theoretical analysis}.
In order to study how pixel rebalance affects $\mathrm{IoU}$,
we examine the connections between $\mathrm{IoU}$ and $\mathrm{Acc}$. We
use $\mathrm{TP}$, $\mathrm{FN}$ and $\mathrm{FP}$ to denote true positives, false negatives, and false positives, respectively. Eq.~\eqref{eq:IoU} presents the formula for $\mathrm{IoU}$, while Eq.~\eqref{eq:Acc} is for Acc. By combining these two equations in Eq.~\eqref{eq:IoU_Acc}, we can see that $\cfrac{1}{\mathrm{IoU}}$ is a linear function of $\cfrac{1}{\mathrm{Acc}}$, with $\cfrac{\mathrm{FP}}{\mathrm{TP}}$ as a constant. In this sense, it is reasonable to optimize $\mathrm{IoU}$ with pixel cross-entropy.
\begin{eqnarray}
\mathrm{IoU} &=& \frac{\mathrm{TP}}{\mathrm{TP}+\mathrm{FN}+\mathrm{FP}}, \label{eq:IoU} \\
\mathrm{Acc} &=& \frac{\mathrm{TP}}{\mathrm{TP}+\mathrm{FN}}, \label{eq:Acc} \\
\frac{1}{\mathrm{IoU}} &=& \frac{1}{\mathrm{Acc}} + \frac{\mathrm{FP}}{\mathrm{TP}}. \label{eq:IoU_Acc}
\end{eqnarray}

When rebalancing with pixel cross-entropy, $\mathrm{Acc}$ of low-frequency classes is expected to improve. However, according to Remark \ref{thm:IoU_Acc}, a higher $\mathrm{Acc}$ does not always mean a higher $\mathrm{IoU}$. Even worse, we empirically observe that Remark \ref{thm:IoU_Acc} is usually not satisfied for pixel rebalancing as shown in the following case study.

\begin{theorem}
	\label{thm:IoU_Acc}
	\normalfont{(Condition for $\mathrm{IoU}$ improvement)~~}
	With pixel rebalance, suppose the $\mathrm{Acc}$ of class $y$ is improved from $A$ to $(1+K)A$, then it must satisfy $\mathrm{FP}^{r}$ - $(1+K) \cdot \mathrm{FP}$ $\leq$ $K \cdot$ $n_{y}$ to guarantee $\mathrm{IoU}$ improvements, where $\mathrm{FP}^{r}$ denotes false positives after the rebalance while $\mathrm{FP}$ is false positives beforehand. $n_{y}$ is the pixel frequency for class $y$. $K \geq 0$.
	\textit{Proof~~} See the Supplementary Material.
\end{theorem} 

\begin{table}[t]
	\centering
	\setlength{\tabcolsep}{22pt}
	\caption{Pixel rebalance with balanced softmax \cite{DBLP:conf/nips/RenYSMZYL20} and re-weighting on ADE$20$K. ResNet-$50$ and Deeplabv$3$+ are adopted. * denotes that region frequency prior is used for rebalance.}
	\label{tab:ade20k_main}
	{
		\begin{tabular}{l|c}
			\shline
			Method &$\mathrm{mIoU}$(s.s.) \\ %&$\mathrm{mAcc}$(s.s.)\\
			\shline
			Baseline                 &$43.95$ \\ %&$54.96$ \\
			\shline
			Re-weighting             &$37.86$ \\%&$62.57$ \\
			balancedsoftmax          &$38.57$ \\ %&$66.07$ \\
			balancedsoftmax*         &$40.66$ \\ %&$63.40$ \\
			%\shline \rowcolor{pearDark!20}
            \hline
			RR (Ours)  &$\bf{45.02}$ \\ %&$55.71$ \\
			\shline
		\end{tabular}
	}
	\label{tab:balancesoftmax_ade20k}
\end{table}

\vspace{1mm}
\noindent\textbf{A Case Study}.
To rebalance with pixel cross-entropy, we conduct experiments on ADE$20$K with balanced softmax \cite{DBLP:conf/nips/RenYSMZYL20}, a state-of-the-art method for long-tailed image classification. As shown in Table \ref{tab:balancesoftmax_ade20k}, the overall $\mathrm{mIoU}$ becomes worse after rebalance, dropping from \textbf{$43.95$} to \textbf{$38.57$}, despite significant improvements in overall $\mathrm{mAcc}$. 

To understand whether the overall $\mathrm{mIoU}$ degradation is caused by over-rebalance, {\textit i.e.}, $\mathrm{IoU}$ of high-frequency classes drops a lot though $\mathrm{IoU}$ of low-frequency classes improves, we plot $\Delta \mathrm{IoU}$ and $\Delta \mathrm{Acc}$ for each class $y$. The $\Delta \mathrm{IoU}$ and $\Delta \mathrm{Acc}$ are calculated by
\begin{eqnarray}
   \Delta \mathrm{IoU}_{y} &=& \mathrm{IoU}_{y}^{\rm{rebalance}} - \mathrm{IoU}_{y}^{\rm{baseline}}, \\
   \Delta \mathrm{Acc}_{y} &=& \mathrm{Acc}_{y}^{\rm{rebalance}} - \mathrm{Acc}_{y}^{\rm{baseline}}.
\end{eqnarray}
As shown in Figure \ref{fig:baseline_vs_pixrebalance} (b) and (c), after pixel rebalance with balanced softmax, $\mathrm{IoU}$ decreases while $\mathrm{Acc}$ increases for low-frequency classes. 

This case study shows that pixel rebalance can improve $\mathrm{Acc}$ while failing to meet the $\mathrm{IoU}$ improvement condition of Remark \ref{thm:IoU_Acc}, which implies that pixel rebalance improves low-frequency $\mathrm{Acc}$ largely through an increase of $\mathrm{FP}$, which is harmful to $\mathrm{IoU}$.

\subsubsection{Accuracy vs. Frequency}
\label{sec:challenge2}
An important prior for rebalancing in long-tailed image classification is that images are nearly i.i.d., which makes class accuracy positively related to the number of images in the corresponding classes as shown in Figure \ref{fig:accs_cifar100_ade20k} (a). 

Under this case, to encourage the optimization process to be more friendly to low-frequency classes in training, it would be sensible to re-weight the cross-entropy loss with a factor that is inversely proportional to class frequency, like in \cite{cb-focal}. Moreover, Cao et al. \cite{cao2019learning} and Ren et al. \cite{DBLP:conf/nips/RenYSMZYL20} theoretically reveal the close relationship between the optimal margin and class frequency, further indicating the importance of class frequency statistics for conducting rebalance.

However, the situation is different for the semantic segmentation task. Pixels within the same region of an image usually belong to the same object or stuff and are thus highly correlated. Since pixels are not i.i.d., the class pixel frequency is an unsuitable factor for rebalancing. This conclusion is supported empirically in Figure \ref{fig:accs_cifar100_ade20k} (b), where class accuracy is shown to have much weaker correlation with class pixel frequency compared to class image frequency in image classification, shown in Figure \ref{fig:accs_cifar100_ade20k} (a).

\begin{table}[t]
	\centering
	\setlength{\tabcolsep}{26pt}
	\caption{Pearson coefficients between class frequency and accuracy.}
	\label{tab:pearson_coefficient}
	{
		\begin{tabular}{l|c}
			\shline
			Dataset &$\rho_{X,Y} (\%)$ \\
			\shline
			CIFAR-$100$       & $75.9$ \\
			ADE$20$K-pixel    & $35.0$ \\
			ADE$20$K-region   & $41.3$ \\
			\shline
		\end{tabular}
	}
	\vspace{-0.1in}
\end{table}

To quantify the correlation between class accuracy and frequency, we calculate the Pearson correlation coefficient \cite{benesty2009pearson}:
\begin{equation}
\rho_{X,Y} = \frac{Cov(X,Y)}{\sigma(X) \sigma(Y)} = \frac{\mathbb{E}[(X-\mu(X))(Y-\mu(Y))]}{\sigma(X) \sigma(Y)},
\end{equation}
where $\mu(.)$, $\sigma(.)$, $Cov(.)$ and $\mathbb{E}(.)$ are functions for mean, standard deviation, covariance and expectation.
Table \ref{tab:pearson_coefficient} lists the Pearson coefficients between class accuracy and image frequency on CIFAR-100, pixel frequency on ADE20K, and region frequency on ADE20K, respectively. The correlations between class accuracy and image frequency, region frequency, pixel frequency decrease step by step. Region frequency has a stronger correlation with class accuracy than pixel frequency, as it eliminates the effects of correlations among neighboring pixels. 
The weak relationship between class accuracy and pixel frequency causes pixel frequency to be less informative for rebalancing, which adds to the challenge of rebalancing in semantic segmentation.

\vspace{-0.1in}
\subsection{Region Rebalance Framework}
Due to the issues discussed in Sec. \ref{sec:challenges_rebalance},
we propose to rebalance at the region level instead of the pixel level, through a \textbf{Region Rebalance (RR)} training strategy. The region rebalance method relieves the class imbalance issue with an auxiliary region classification branch by encouraging features to lie in a more balanced region classification space. Moreover, rebalance methods from long-tailed image classification can be readily employed to solve the region imbalance problem. This involves no extra cost for testing because the region rebalance branch is removed during inference.

The region rebalance framework shown in Figure \ref{fig:RRT_method} contains two components: 
(i) An auxiliary region classification branch to ease the class imbalance problem and (ii) Enhanced rebalance with techniques in long-tailed image classification for solving region imbalance. The overall loss functions in the following are deployed:

\vspace{-0.1in}
\begin{eqnarray}
   \mathcal{F}(i) \!&=&\! \sum_{x \in \mathcal{D}} 1 \!\quad\! if \!\quad\! i \in x_{gt} \!\quad\! otherwise \!\quad\! 0, \\
   \mathcal{L}_{reg} \!&=&\! \frac{1}{|\mathcal{R}|} \sum_{r \in \mathcal{R}} -\log(\frac{\mathcal{F}(r^{y}) e^{r_{y}} }{\sum_{k \in Y} \mathcal{F}(k) e^{r_{k}}}) \label{eq:region},\\
   \mathcal{L}_{all} \!&=&\! \mathcal{L}_{pixel} + \lambda \mathcal{L}_{reg} \label{eq:all},
\end{eqnarray}
where $\mathcal{D}$ is the dataset, $\mathcal{R}$ contains all the extracted regions, $r^{y}$ is the ground truth label of region $r$, $r_{i}$ represents the region logit corresponding to class $i$, $\mathcal{F}(i)$ denotes the region frequency of class $i$, $Y$ is the label space, $\mathcal{L}_{reg}$ is the region loss, while $\mathcal{L}_{pixel}$ represents pixel cross-entropy loss.

In the region classification branch, we group pixels belonging to the same class into region features by averaging the pixel features according to the ground-truth region map. Then the extracted region features are fed into a region classifier.
Further, we enhance the rebalance using a technique \cite{DBLP:conf/nips/RenYSMZYL20} borrowed from the long-tailed image classification community to tackle the region imbalance problem. The deployed region loss is shown in Eq.~\eqref{eq:region}, and we use a hyper-parameter $\lambda$ to control the strength of rebalance, shown in Eq.~\eqref{eq:all}.

\subsection{Analysis of Our Method}
The misalignment between the training objective, {\textit i.e.}, pixel cross-entropy, and evaluation metric $\mathrm{IoU}$ at inference is problematic for direct pixel rebalance as analyzed in Sec.~\ref{sec:challenges_rebalance}. The proposed method instead relieves class imbalance via region rebalance. To clarify the mechanism of our region rebalance method, we analyze it in comparison to pixel rebalance. In addition, we discuss differences in imbalance reflected by dataset statistics.

\begin{figure*} 
\begin{center}
\begin{minipage}[t]{0.66\textwidth}
\footnotesize
\begin{center}
\begin{subfigure}[b]{0.5\linewidth}
\centering
\hspace{-12.5mm}
\includegraphics[width=1\textwidth]{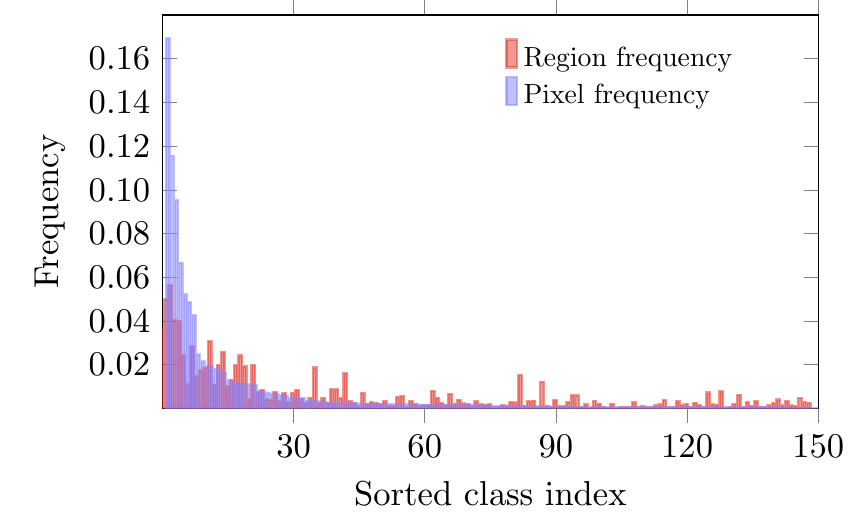}
\end{subfigure}%
\begin{subfigure}[b]{0.5\linewidth}
\centering
\hspace{-12mm}
\includegraphics[width=1\textwidth]{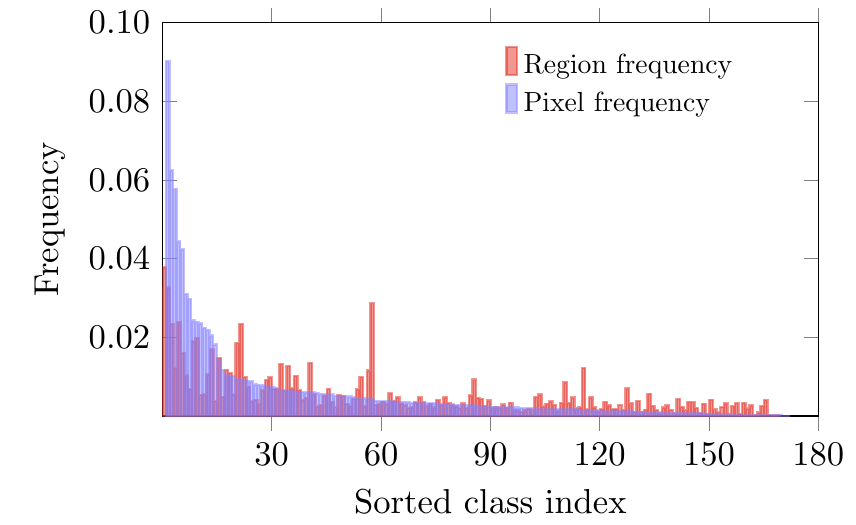}
\end{subfigure}
\subcaption{long-tailed distribution of pixels vs. regions}
\end{center}
\end{minipage}
\begin{minipage}[t]{0.3\textwidth} 
\vspace{-1.2in}
\centering
\footnotesize
\setlength{\tabcolsep}{10pt}
\renewcommand{\arraystretch}{1.8}
\subcaption{degree of pixel imbalance vs. region imbalance.}
%\vspace{0.2in}
{
\begin{tabular}{l|cc}
\shline
Dataset &PIF &RIF \\
\shline
ADE$20$K          &$827$  &$282$ \\
COCO-Stuff$164$K  &$2612$ &$528$ \\
\shline
\end{tabular}
}
\label{tab:statistics_imbalance}
\end{minipage}
\end{center}
\vspace{-0.22in}
\caption{
\textbf{Comparison between pixel imbalance and region imbalance:}
(a) Histogram statistics of pixel/region frequency over sorted class indexes on ADE$20$K (left) and COCO-Stuff$164$K (right).
(b) Numerical comparisons of
pixel imbalance factor (PIF) and region imbalance factor (RIF). It can be seen that there is less region imbalance than pixel imbalance.}
\label{fig:data_imbalance_seg}
\vspace{-0.15in}
\end{figure*}

\vspace{-0.1in}
\subsubsection{Region Rebalance vs. Pixel Rebalance}
\label{sec:region_rebalance_analysis}
On one hand, region rebalance gathers pixel features of the same class into region features, and then region classification follows. The ``gather" operation essentially removes the effect of correlation among pixels in a region. As demonstrated in Table~\ref{tab:pearson_coefficient}, region frequency has a stronger correlation with class accuracy than pixel frequency, further indicating its greater suitability for rebalancing. Besides, the ``gather" operation reduces data class imbalance and thus is good for rebalancing as analyzed in Sec. \ref{sec:datasets_stats}.

On the other hand, we empirically observe that region rebalance is more aligned with the $\mathrm{IoU}$ metric. As studied in Sec. \ref{sec:challenge1}, pixel rebalance promotes higher $\mathrm{Acc}$ while failing to improve $\mathrm{IoU}$ because of an increase in $\mathrm{FP}$. 

Such behavior can be understood by considering the situation that two very different objects/stuff can still have similar local pixel features. With pixel rebalance, we attach more importance to pixels of low-frequency classes, and in consequence, corresponding similar pixels in other classes are more likely to become $\mathrm{FP}$. In contrast, objects/stuff are more distinguishable at the region level due to features that are more global. Though more importance is also attached to regions of low-frequency classes with region rebalance, it will be much harder to predict other class objects/stuff as $\mathrm{FP}$. More visual evidence is discussed in our supplementary file. We leave more theoretical analysis to future work.

\vspace{-0.12in}
\subsubsection{Datasets Statistics}
\label{sec:datasets_stats}
Semantic segmentation datasets suffer from both pixel level and region level data imbalance. However, we observe that the degree of pixel imbalance is more serious than that of region imbalance as shown in Figure~\ref{fig:data_imbalance_seg}. We collect statistics from the most popular semantic segmentation benchmarks in Figure \ref{tab:statistics_imbalance}. Following convention \cite{cb-focal,liu2019large}, we calculate the imbalance factor (IF) as $\beta= \frac{N_{max}}{N_{min}}$, where
$N_{max}$ and $N_{min}$ are the numbers of training samples for the most frequent class and the least frequent class.

As shown in Figure \ref{tab:statistics_imbalance}, for ADE$20$K, the pixel imbalance factor (PIF) is nearly $3$ times the region imbalance factor (RIF). For COCO-Stuff$164$K, PIF is nearly $5$ times RIF. 
As our method relieves the class imbalance issue via region rebalancing, it conveniently benefits from less imbalance than pixel rebalance would, and this in turn promotes more effective rebalancing.

\section{Experiments}
To evaluate the effectiveness of our method, we conduct experiments on the most popular benchmarks in semantic segmentation, {\textit i.e.}, ADE$20$K \cite{zhou2017scene} and COCO-Stuff$164$K \cite{caesar2018coco}. By inserting the proposed region rebalance method into UperNet, PSPNet, Deeplabv$3$+, and OCRNet, clear improvements are obtained. We also test the region rebalance method on popular transformer neural networks, specifically Swin transformer \cite{liu2021swin} and \textsc{BEiT} \cite{bao2021beit}.

\subsection{Datasets}
\vspace{1mm}
\noindent\textbf{ADE$20$K}.
ADE$20$K \cite{zhou2017scene} is a challenging dataset often used to validate transformer-based neural networks on downstream tasks such as semantic segmentation. It contains $22$K densely annotated images with 150 fine-grained semantic concepts. The training and validation sets consist of $20$K and $2$K images, respectively.

\vspace{1mm}
\noindent\textbf{COCO-Stuff$164$K}.
COCO-Stuff$164$K \cite{caesar2018coco} is a large-scale scene understanding benchmark that can be used for evaluating semantic segmentation, object detection, and image captioning. It includes all $164$K images from COCO $2017$. The training and validation sets contain $118$K and $5$K images, respectively. It covers $171$ classes: $80$ thing classes and $91$ stuff classes.

\subsection{Implementation Details}
We implement the proposed region rebalance method in the mmsegmentation codebase \cite{mmseg2020} and follow the commonly used training settings for each dataset. More details are described in the following.

For backbones, we use CNN-based ResNet-$50$c and ResNet-$101$c, which replace the first $7 \times 7$ convolution layer in the original ResNet-$50$ and ResNet-$101$ with three consecutive $3 \times 3$ convolutions. Both are popular in the semantic segmentation community \cite{DBLP:conf/cvpr/ZhaoSQWJ17}. For OCRNet, we adopt HRNet-W$18$ and HRNet-W$48$ \cite{wang2020deep}. For transformer-based neural networks, we adopt the popular Swin transformer \cite{liu2021swin} and \textsc{BEiT} \cite{bao2021beit}. \textsc{BEiT} achieves the most recent state-of-the-art performance on the ADE$20$K validation set. "$\dag$" in Tables \ref{tab:main_ade20k} and \ref{tab:main_cocostuff164k} indicates models that are pretrained on ImageNet-$22$K.

With CNN-based models, we use SGD and the poly learning rate schedule \cite{DBLP:conf/cvpr/ZhaoSQWJ17} with an initial learning rate of $1e-2$ and a weight decay of $1e-4$. If not stated otherwise, we use a crop size of $512 \times 512$, a batch size of $16$, and train all models for $160$K iterations on ADE$20$K and $320$K iterations on COCO-Stuff164K. For Swin transformer and \textsc{BEiT}, we use their default optimizer, learning rate setup and training schedule.

In the training phase, the standard random scale jittering between $0.5$ and $2.0$, random horizontal flipping, random cropping, as well as random color jittering are used as data augmentations \cite{mmseg2020}. For inference, we report the performance of both single scale (s.s.) inference and multi-scale (m.s.) inference with horizontal flips and
scales of $0.5$, $0.75$, $1.0$, $1.25$, $1.5$, $1.75$.

\begin{table}[t]
	\centering
	\setlength{\tabcolsep}{2.0pt}
	\caption{Performance on ADE$20$K.}
	\resizebox{1.0\linewidth}{!}
	{
		\begin{tabular}{l|ccccc}
			\shline
			Method &Backbone  &$\mathrm{mIoU}$(s.s.) &$\mathrm{mIoU}$(m.s.) \\
			\shline
			OCRNet              &HRNet-W$18$   &$39.32$ &$40.80$ \\
			OCRNet-RR           &HRNet-W$18$   &$41.80$ &$43.55$($\textcolor{mylight}{+2.75}$) \\
			UPerNet             &ResNet-$50$  &$42.05$ &$42.78$ \\
			UPerNet-RR          &ResNet-$50$  &$43.27$ &$44.03$($\textcolor{mylight}{+1.25}$) \\
			PSPNet              &ResNet-$50$  &$42.48$ &$43.44$ \\ 
			PSPNet-RR           &ResNet-$50$  &$42.83$ &$44.26$($\textcolor{mylight}{+0.82}$) \\ 
			DeepLabv$3$+        &ResNet-$50$  &$43.95$ &$44.93$ \\
			DeepLabv$3$+-RR     &ResNet-$50$  &$45.02$ &$46.03$($\textcolor{mylight}{+1.10}$) \\
			\shline
			OCRNet              &HRNet-W$48$   &$43.25$ &$44.88$ \\
			OCRNet-RR           &HRNet-W$48$   &$44.50$ &$46.09$($\textcolor{mylight}{+1.21}$) \\
			UperNet             &ResNet-$101$ &$43.82$ &$44.85$ \\
			UperNet-RR          &ResNet-$101$ &$44.78$ &$46.11$($\textcolor{mylight}{+1.26}$) \\
			PSPNet              &ResNet-$101$ &$44.39$ &$45.35$ \\
			PSPNet-RR           &ResNet-$101$ &$44.65$ &$46.15$($\textcolor{mylight}{+0.80}$) \\
            DeepLabv$3$+        &ResNet-$101$ &$45.47$ &$46.35$ \\
			DeepLabv$3$+-RR     &ResNet-$101$ &$46.67$ &$47.96$($\textcolor{mylight}{+1.61}$) \\
			\shline
			UperNet       &Swin-T       &$44.51$ &$45.81$ \\
			UperNet-RR    &Swin-T       &$45.02$ &$46.45$($\textcolor{mylight}{+0.64}$) \\
			UperNet       &Swin-B\dag   &$50.04$ &$51.66$ \\
			UperNet-RR    &Swin-B\dag   &$51.20$ &$52.88$($\textcolor{mylight}{+1.22}$) \\
			\shline
			UperNet       &\textsc{BEiT}-L &$56.7$ &$57.0$ \\
			UperNet-RR    &\textsc{BEiT}-L &$57.2$ &$57.7$($\textcolor{mylight}{+0.70}$) \\
			\shline
		\end{tabular}
	}
	\label{tab:main_ade20k}
	\vspace{-0.2in}
\end{table}

\subsection{Main Results}

\vspace{1mm}
\noindent\textbf{Comparison on ADE$20$K}.
To show the flexibility of our region rebalance module, we experiment on ADE$20$K with various semantic segmentation methods, {\textit e.g.}, PSPNet, UperNet, OCRNet, and Deeplabv3+. As shown in Table \ref{tab:main_ade20k}, plugging the region rebalance module into those methods leads to significant improvements. Specifically, for ResNet-$101$ or HRNet-W$48$, after training with Region Rebalance, there are $1.26$\%, $0.80$\%, $1.21$\% and $1.61$\% gains for UperNet, PSPNet, OCRNet, and Deeplabv3+ respectively.

To further show the effectiveness of the proposed region rebalance method, we experiment with Swin transformer and \textsc{BEiT}. For CNN-based models, we achieve $47.96$ $\mathrm{mIoU}$ with ResNet-$101$, surpassing the baseline by $1.61$. With \textsc{BEiT}, the state-of-the-art is improved from $57.0$ $\mathrm{mIoU}$ to $57.7$ $\mathrm{mIoU}$.

\vspace{1mm}
\noindent\textbf{Comparison on COCO-Stuff$164$K}.
On the large-scale COCO-Stuff$164$K dataset, we again demonstrate the flexibility of our region rebalance module. The experimental results are summarized in Table~\ref{tab:main_cocostuff164k}.
Equipped with the region rebalance module in training, CNN-based models, {\textit i.e.}, ResNet-$50$, HRNet-W$18$, ResNet-$101$ and HRNet-W$48$, surpass their baselines by a large margin. Specifically, with HRNet-$18$ and OCRNet, our trained model outperforms the baseline by $6.44$ $\mathrm{mIoU}$. With the large CNN-based ResNet-$101$ and Deeplabv3+, our model achieves $44.62$ $\mathrm{mIoU}$, surpassing the baseline by $1.66$ $\mathrm{mIoU}$. 

We also verify the effectiveness of the region rebalance module with Swin transformer on COCO-Stuff$164$K. Experimental results with Swin-T and Swin-B show clear improvements after rebalancing with our region rebalance method in the training phase.

\begin{table}[t]
	\centering
	\setlength{\tabcolsep}{2.0pt}
	\caption{Performance on COCOStuff-$164$K.}
	\resizebox{1.00\linewidth}{!}
	{
		\begin{tabular}{l|ccc}
			\shline
			Method &Backbone  &$\mathrm{mIoU}$(s.s.) &$\mathrm{mIoU}$(m.s.) \\
			\shline
			OCRNet               &HRNet-W$18$   &$31.58$ &$32.34$ \\
			OCRNet-RR            &HRNet-W$18$   &$37.98$ &$38.78$($\textcolor{mylight}{+6.44}$) \\
			UperNet              &ResNet-$50$  &$39.86$ &$40.26$ \\
			UperNet-RR           &ResNet-$50$  &$41.20$ &$41.73$($\textcolor{mylight}{+1.47}$) \\
			PSPNet               &ResNet-$50$  &$40.53$ &$40.75$ \\
			PSPNet-RR            &ResNet-$50$  &$41.92$ &$42.34$($\textcolor{mylight}{+1.95}$) \\
			DeepLabv$3$+         &ResNet-$50$  &$40.85$ &$41.49$                \\
			DeepLabv$3$+-RR      &ResNet-$50$  &$42.66$ &$43.47$($\textcolor{mylight}{+1.89}$) \\
			\shline
			OCRNet               &HRNet-W$48$    &$40.40$ &$41.66$ \\
			OCRNet-RR            &HRNet-W$48$    &$41.83$ &$43.29$($\textcolor{mylight}{+1.36}$)\\
			UperNet              &ResNet-$101$  &$41.15$ &$41.51$ \\
			UperNet-RR           &ResNet-$101$  &$42.30$ &$42.78$($\textcolor{mylight}{+1.27}$) \\
			PSPNet               &ResNet-$101$  &$41.95$ &$42.42$ \\
			PSPNet-RR            &ResNet-$101$  &$43.40$ &$43.85$($\textcolor{mylight}{+1.43}$) \\
            DeepLabv$3$+         &ResNet-$101$  &$42.39$ &$42.96$\\
			DeepLabv$3$+-RR      &ResNet-$101$  &$43.93$ &$44.62$($\textcolor{mylight}{+1.66}$) \\
			\shline
			UperNet              &Swin-T         &$43.83$ &$44.58$ \\
			UperNet-RR           &Swin-T         &$44.45$ &$45.18$($\textcolor{mylight}{+0.60}$) \\
			UperNet              &Swin-B\dag     &$47.67$ &$48.57$ \\
			UperNet-RR           &Swin-B\dag     &$48.21$ &$49.20$($\textcolor{mylight}{+0.63}$) \\
			\shline
		\end{tabular}
	}
    \label{tab:main_cocostuff164k}
	\vspace{-0.2in}
\end{table}

\begin{figure*}[t]
	\begin{center}
		\begin{minipage}[t]{0.32\textwidth} 
			\begin{center}
				\includegraphics[width=0.99\textwidth]{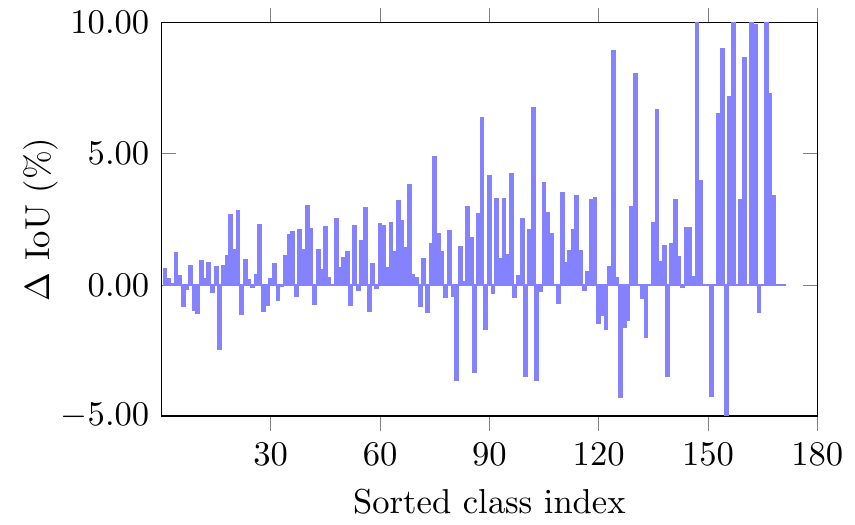} 
				\subcaption{Deeplabv$3$+ w/ ResNet-$101$.} 
				\label{fig:residual_iou_deeplabv3plus} 
			\end{center} 
		\end{minipage} 
		\hspace{0.05in}
		\begin{minipage}[t]{0.32\textwidth} 
			\begin{center}
				\includegraphics[width=0.99\textwidth]{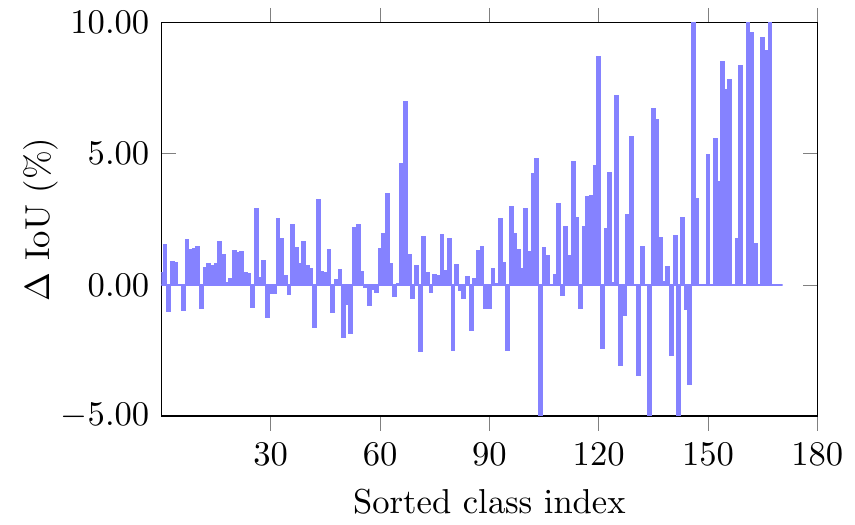} 
				\subcaption{OCRNet w/ HRNet-W$48$.} 
				\label{fig:residual_iou_ocrnet} 
			\end{center} 
		\end{minipage} 
		\hspace{0.05in}
		\begin{minipage}[t]{0.32\textwidth} 
			\begin{center}
				\includegraphics[width=0.99\textwidth]{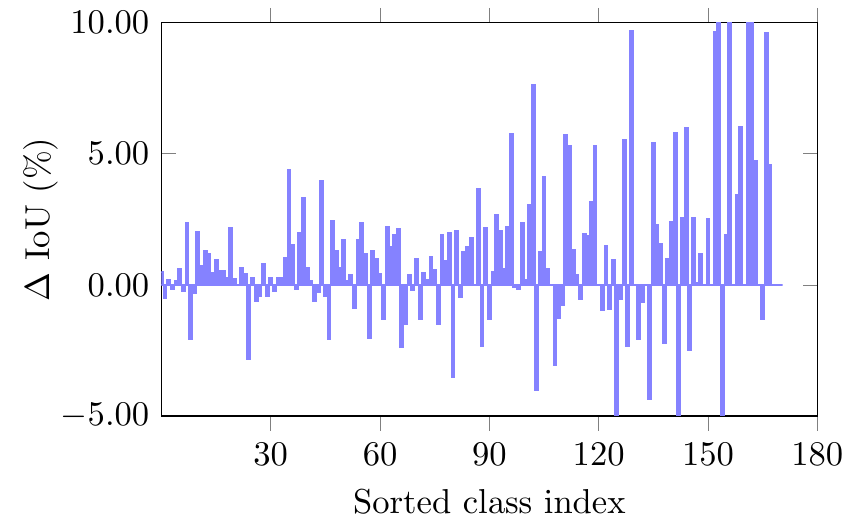} 
				\subcaption{UperNet w/ ResNet-$101$.} 
				\label{fig:residual_iou_upernet} 
			\end{center}
		\end{minipage}
	\end{center}
	\vspace{-0.15in}
	\caption{\textbf{Illustrating the category-wise improvements with our region rebalance method}: The $\mathrm{IoU}$s of most low-frequency classes are improved when the proposed region rebalance method is adopted in training. Class indexes are sorted in descending order by the number of pixels belonging to the same class.}
	\label{fig:residual_iou_cocostuff164k}
	\vspace{-0.1in}
\end{figure*}

\subsection{Ablation Experiments}

\begin{figure}[t]
\centering
\includegraphics[width=0.4\textwidth]{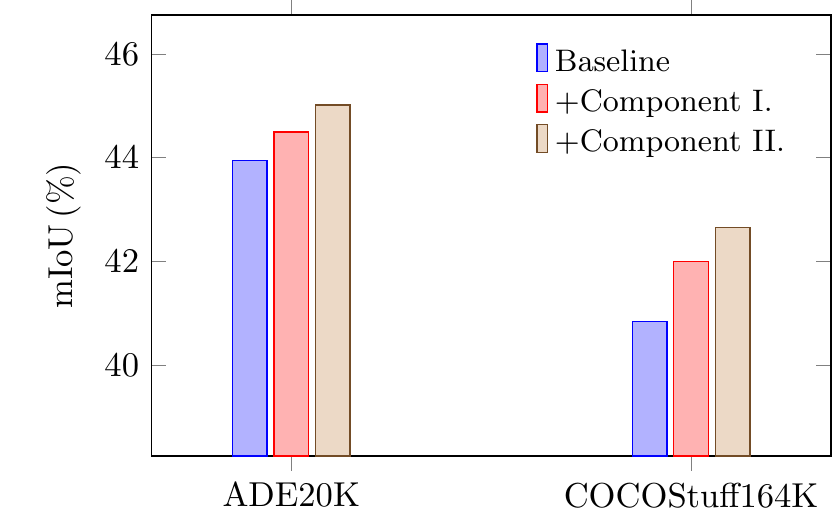}
\caption{Ablations for our region rebalance components.}
\label{fig:ablation_RRT}
\vspace{-0.1in}
\end{figure}

\vspace{1mm}
\noindent\textbf{Ablation of region rebalance components}.
To verify the usefulness of both components in our region rebalance method, we conduct the following ablations on top of the baseline: I. only with an auxiliary region classification branch; II. further inserting a rebalance mechanism, {\textit i.e.}, balanced softmax \cite{DBLP:conf/nips/RenYSMZYL20}. We plot the experimental results in Figure \ref{fig:ablation_RRT}.

With the two components applied in training, we observe consistent improvements on ADE$20$K and the large-scale COCO-Stuff$164$K. Additionally, an interesting phenomenon is observed: with only component I, deep models already enjoy significant gains on COCO-Stuff$164$K. The reason behind this may be that there is a much larger ratio between the pixel imbalance factor (PIF) and region imbalance factor (RIF) on COCO-Stuff$164$K than on ADE$20$K, as shown in Table \ref{tab:statistics_imbalance}. From this point of view, when adopting I, models are given stronger regularization and the output features are encouraged to be more balanced.

\vspace{1mm}
\noindent\textbf{Improvements on low-frequency classes}.
To further examine the rebalancing effects of our proposed region rebalance method, we plot the class $\Delta \mathrm{IoU}$ between the baseline model and the model trained with the region rebalance method. Figure \ref{fig:baseline_vs_pixrebalance} (a) shows the results on ADE$20$K with ResNet-$50$ and Deeplabv$3$+. We also validate the rebalance effects on COCO-Stuff$164$K with different semantic segmentation methods, {\textit i.e.}, Deeplabv$3$+, OCRNet, and UperNet, as shown in Figure~\ref{fig:residual_iou_cocostuff164k}. We observe that the $\mathrm{IoU}$ of most low-frequency classes is enhanced. For example, the $\mathrm{IoU}$ of class ``shower" is improved from $0$ to $6.14$ in ADE$20$K.

\begin{table}[t]
	\centering
	\setlength{\tabcolsep}{24pt}
	\caption{Ablation on hyper-parameter $\lambda$.}
	\label{tab:ablation_lamda}
	{
		\begin{tabular}{l|c}
			\shline
			Hyper-parameter $\lambda$  &$\mathrm{mIoU}$(s.s.) \\
			\shline
			$\lambda=0.0$ (baseline) &43.95 \\
			\shline
            $\lambda=0.1$  &44.61\\
            $\lambda=0.2$  &44.85\\
            $\lambda=0.3$  &45.02\\
            $\lambda=0.4$  &44.64\\
			\shline
		\end{tabular}
	}
	\vspace{-0.15in}
\end{table}

\vspace{1mm}
\noindent\textbf{Ablation of hyper-parameter $\lambda$}.
In Eq.~\eqref{eq:all}, we introduce the hyper-parameter $\lambda$ for weighting between pixel loss and region loss. All experimental results in Tables~\ref{tab:main_ade20k} and \ref{tab:main_cocostuff164k} are reported for the same $\lambda=0.3$. 

To show the sensitivity of our region rebalance method to different $\lambda$ values, we conduct an ablation with ResNet-50 and Deeplabv3+ on ADE20K. Table \ref{tab:ablation_lamda} lists the experimental results, showing that the performance is not affected substantially by the value of $\lambda$ within the range [0.1,0.3].

\vspace{1mm}
\noindent\textbf{Ablation on Dice loss}.
Though some $\mathrm{IoU}$-based loss functions, {\textit e.g.}, Dice Loss \cite{milletari2016v}, have been developed, we empirically observe that pixel cross-entropy is still necessary in training for high performance. An ablation on the Dice loss is conducted on ADE$20$K with ResNet-50 and Deeplabv$3$+, and the results are reported in Table \ref{tab:ablation_dice}. With just the Dice Loss itself, the model achieves only 1.14 mIoU. This huge performance degradation implies that the pixel cross-entropy is still a necessary part and the challenges in Sec. \ref{sec:challenges_rebalance} will still exist when we conduct pixel rebalance.

\begin{table}[t]
	\centering
	\setlength{\tabcolsep}{15pt}
	\caption{Ablation on Dice loss.}
	\label{tab:ablation_dice}
	{
		\begin{tabular}{l|c}
			\shline
			Method  &$\mathrm{mIoU}$(s.s.) \\
			\shline
			Baseline (cross-entropy)    &43.95 \\
			\shline
            Dice Loss                   &1.14 \\
            Dice Loss + cross-entropy   &44.12\\
			\shline
			RR (Ours)                   &45.02\\
			\shline
		\end{tabular}
	}
	\vspace{-0.15in}
\end{table}

\vspace{-0.1in}
\section{Conclusion}
\vspace{-0.1in}
In this paper, we investigate the imbalance phenomenon in semantic segmentation and identified two major challenges in conducting pixel rebalance, {i.e.}, (i) misalignment between the training objective and the $\mathrm{IoU}$ evaluation metric at inference, and (ii) correlations among pixels within the same object/stuff cause class frequency to be less effective for rebalancing. With our analysis, we propose to gather correlated pixels into regions and rebalance within the context of region classification. The proposed region rebalance method leads to strong improvements across different semantic segmentation methods, {\textit e.g.}, OCRNet, DeepLabv$3$+, UperNet, and PSPNet. Experimental results on ADE$20$K and the large-scale COCO-Stuff$164$K verify the effectiveness of our method.

{\small
\bibliographystyle{ieee_fullname}
\bibliography{egbib}
}

\newpage
\onecolumn
\appendix

\begin{center}
	\Large \textbf{Region Rebalance for Long-Tailed Semantic Segmentation}
	\Large \\ \textbf{Supplementary Material}
\end{center}
\vspace{20pt}

\section{Experiments on COCO-Stuff$10$K for Semantic Segmentation}
\noindent COCO-Stuff$10$K, which includes $10$K images from the COCO training set, is a subset of COCO-Stuff$164$K. The training and validation sets consist of $9$K and $1$K images, respectively. It also covers $171$ classes. The experimental results are reported in Table \ref{tab:main_cocostuff10k}.
Employing our region rebalance module in training yields $1.18$ $\mathrm{mIoU}$ and $1.19$ $\mathrm{mIoU}$ improvements separately when compared with the baselines.

\begin{table}[h]
	\centering
	\setlength{\tabcolsep}{20.0pt}
	\caption{Performance on COCO-Stuff$10$K.}
	\label{tab:main_cocostuff10k}
	%\resizebox{1.00\linewidth}{!}
	{
		\begin{tabular}{l|ccccc}
			\shline
			Method &Backbone  &$\mathrm{mIoU}$(s.s.) &$\mathrm{mIoU}$(m.s.) \\
			\shline
			DeepLabv$3$+         &ResNet-$50$  &$35.84$ &$36.85$                 \\
			DeepLabv$3$+-RR      &ResNet-$50$  &$36.53$ &$38.03$($\textcolor{mylight}{+1.18}$) \\
            DeepLabv$3$+         &ResNet-$101$ &$38.14$ &$39.00$                 \\
			DeepLabv$3$+-RR      &ResNet-$101$ &$38.72$ &$40.19$($\textcolor{mylight}{+1.19}$) \\
			\shline
		\end{tabular}
	}
\end{table}

\vspace{0.5in}
\section{Proof of Remark \ref{thm:IoU_Acc}}
\noindent Before the pixel rebalance, for one low-frequency class $y$, we suppose its $\mathrm{Acc}$ and $\mathrm{IoU}$ are
\begin{eqnarray}
\mathrm{Acc} &=& \frac{\mathrm{TP}}{\mathrm{TP}+\mathrm{FN}},  \\
\mathrm{IoU} &=& \frac{\mathrm{TP}}{\mathrm{TP}+\mathrm{FN}+\mathrm{FP}}. 
\end{eqnarray}
With the pixel rebalance, we expect the $\mathrm{Acc}$ is improved to $\mathrm{Acc} \cdot (1+K)$. Then the class $\mathrm{Acc}$ and $\mathrm{IoU}$ become
\begin{eqnarray}
\mathrm{Acc}^{r} &=& \frac{\mathrm{TP} \cdot (1+K)}{\mathrm{TP} \cdot (1+K) + \mathrm{FN} - K \cdot \mathrm{TP}} = \frac{\mathrm{TP} \cdot (1+K)}{\mathrm{TP}+\mathrm{FN}},  \\
\mathrm{IoU}^{r} &=& \frac{\mathrm{TP} \cdot (1+K)}{\mathrm{TP} \cdot (1+K) + \mathrm{FN} - K \cdot \mathrm{TP} +\mathrm{FP}^{r}} = \frac{\mathrm{TP} \cdot (1+K)}{\mathrm{TP}+\mathrm{FN} + \mathrm{FP}^{r}}. 
\end{eqnarray}
Taking the reciprocal of $\mathrm{IoU}^{r}$ and $\mathrm{IoU}$, we obtain
\begin{eqnarray}
\frac{1}{\mathrm{IoU}^{r}} &=& \frac{\mathrm{TP} \cdot (1+K) + \mathrm{FN} - K \cdot \mathrm{TP} +\mathrm{FP}^{r}}{\mathrm{TP} \cdot (1+K)} = 1 + \frac{\mathrm{FN} - K \cdot \mathrm{TP} + \mathrm{FP}^{r}}{\mathrm{TP} \cdot (1+K)}, \\
\frac{1}{\mathrm{IoU}}     &=& \frac{\mathrm{TP} + \mathrm{FN} + \mathrm{FP}}{\mathrm{TP}}=1 + \frac{\mathrm{FN} + \mathrm{FP}}{\mathrm{TP}} = 1 + \frac{(\mathrm{FN} + \mathrm{FP}) \cdot (1+K)}{\mathrm{TP} \cdot (1+K)}.
\end{eqnarray}
To guarantee $\mathrm{IoU} \leq \mathrm{IoU}^{r}$, it should satisfy:
\begin{eqnarray}
\mathrm{FN} - K \cdot \mathrm{TP} + \mathrm{FP}^{r} \leq (FN+FP) \cdot (1+K) \iff \mathrm{FP}^{r} - (1+K) \cdot \mathrm{FP} \leq K \cdot (\mathrm{TP} + \mathrm{FN})=K \cdot n_{y},
\end{eqnarray}
where $n_{y}$ is the pixel frequency of class $y$.

\newpage
\begin{figure*}[h]
	\begin{center}
		\begin{minipage}[t]{0.48\textwidth} 
			\begin{center}
				\includegraphics[width=0.99\textwidth]{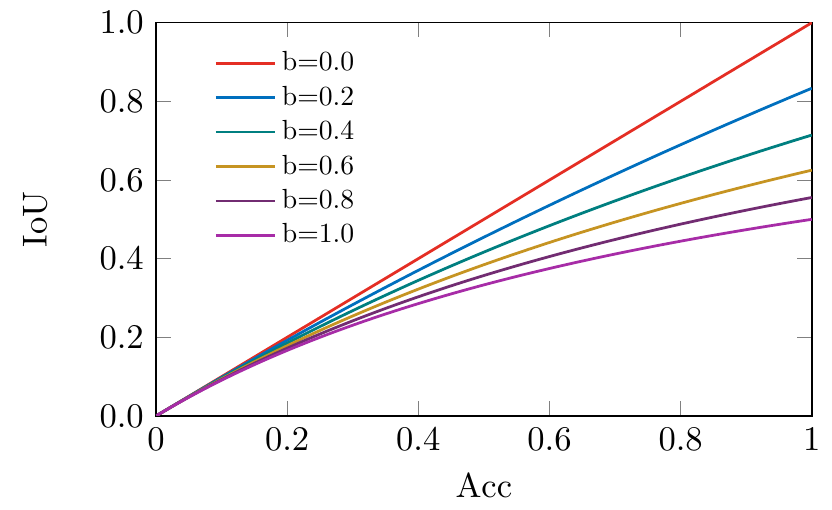} 
				\subcaption{The effects of $\cfrac{\mathrm{FP}}{\mathrm{TP}}$ on $\mathrm{IoU}$ and $\mathrm{Acc}$. with $b=\cfrac{\mathrm{FP}}{\mathrm{TP}} \in [0,1]$.} 
				\label{fig:iou_acc_fp_tp_01} 
			\end{center} 
		\end{minipage} 
		\hspace{0.05in}
		\begin{minipage}[t]{0.48\textwidth} 
			\begin{center}
				\includegraphics[width=0.99\textwidth]{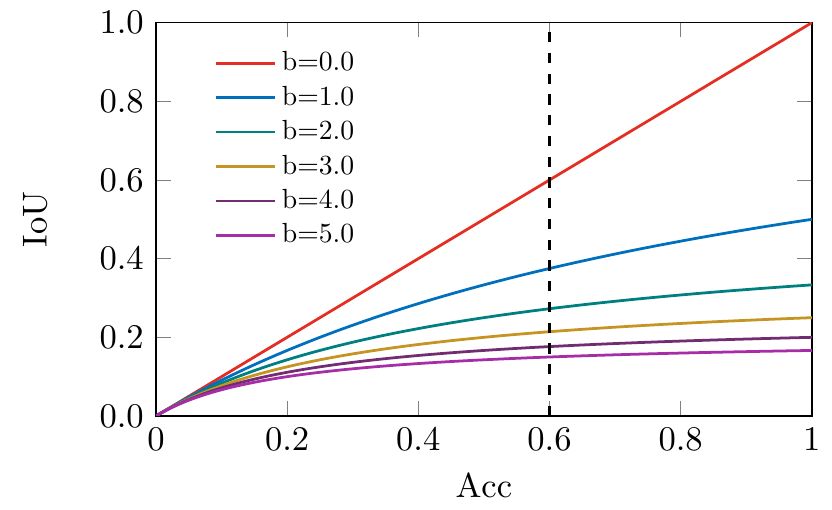} 
				\subcaption{The effects of $\cfrac{\mathrm{FP}}{\mathrm{TP}}$ on $\mathrm{IoU}$ and $\mathrm{Acc}$. with $b=\cfrac{\mathrm{FP}}{\mathrm{TP}} \in [0,5]$.} 
				\label{fig:iou_acc_fp_tp_05} 
			\end{center}
		\end{minipage}
	\end{center}
	\caption{The effects of $\cfrac{\mathrm{FP}}{\mathrm{TP}}$ on $\mathrm{IoU}$ and $\mathrm{Acc}$.}
	\label{fig:iou_acc_fp_tp}
\end{figure*}

\section{How Does $\cfrac{\mathrm{FP}}{\mathrm{TP}}$ Make Effects on $\mathrm{IoU}$ and $\mathrm{Acc}$?}
\noindent As analyzed in Sec. \ref{sec:challenges_rebalance}, we examine the connections between $\mathrm{IoU}$ and $\mathrm{Acc}$, {\textit i.e.},
\begin{eqnarray}
\frac{1}{\mathrm{IoU}} = \frac{1}{\mathrm{Acc}} + \frac{\mathrm{FP}}{\mathrm{TP}}.
\end{eqnarray}
To understand the role of $\cfrac{\mathrm{FP}}{\mathrm{TP}}$, we plot the curves with respect to $\mathrm{IoU}$ and $\mathrm{Acc}$ with various constant value $b=\cfrac{\mathrm{FP}}{\mathrm{TP}}$ shown in Figure \ref{fig:iou_acc_fp_tp}. We observe that, when $b=\cfrac{\mathrm{FP}}{\mathrm{TP}}$ is a smaller value around 0.0, {\textit e.g.}, $b \in [0,1]$, optimizing $\mathrm{Acc}$ is equal to optimizing $\mathrm{IoU}$ (Figure \ref{fig:iou_acc_fp_tp_01}). However, demonstrated by Figure \ref{fig:iou_acc_fp_tp_05}, when $b=\cfrac{\mathrm{FP}}{\mathrm{TP}}$ is a larger value, {\textit e.g.}, $b \ge 3$, optimizing $\mathrm{Acc}$ will have little effect on $\mathrm{IoU}$, especially when $\mathrm{Acc}$ has already achieved a high value, {\textit e.g.}, $\mathrm{Acc}=0.6$. This is just the reason that $\mathrm{Acc}$ is improved while $\mathrm{IoU}$ even decrease for low-frequency classes compared to the baseline in Sec. \ref{sec:challenge1}.

\vspace{0.5in}
\section{Visual Evidence for Region Rebalance vs. Pixel Rebalance}
\begin{figure*}[h]
	\centering
	\resizebox{1.0\linewidth}{!}{
	\begin{tabular}{@{\hspace{0.5mm}}c@{\hspace{0.5mm}}c@{\hspace{0.5mm}}c@{\hspace{0.5mm}}c@{\hspace{0.5mm}}c@{\hspace{0.5mm}}c@{\hspace{0.0mm}}}
		\rotatebox[origin=c]{90}{\scriptsize{(a) Input}} 
	    \includegraphics[width=0.16\linewidth,height=0.8in,valign=m]{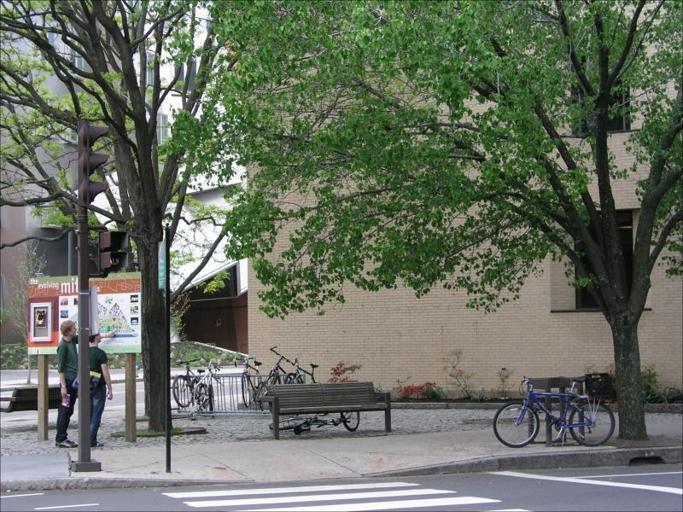}&
		\includegraphics[width=0.16\linewidth,height=0.8in,valign=m]{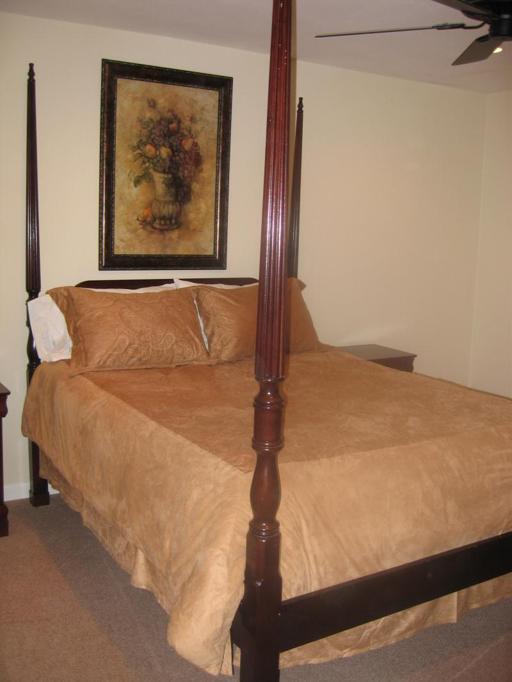}&
		\includegraphics[width=0.16\linewidth,height=0.8in,valign=m]{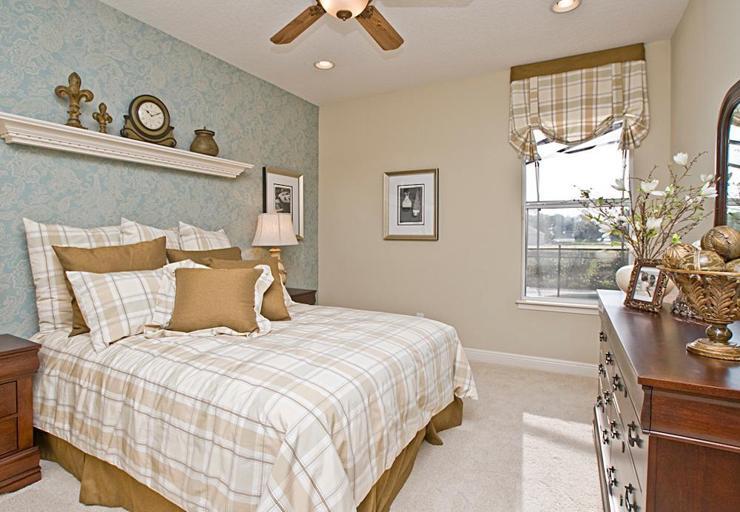}&
		\includegraphics[width=0.16\linewidth,height=0.8in,valign=m]{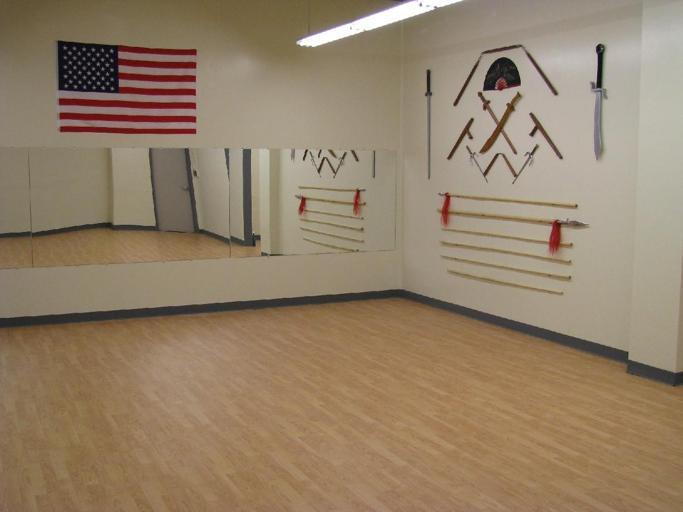}&
		\includegraphics[width=0.16\linewidth,height=0.8in,valign=m]{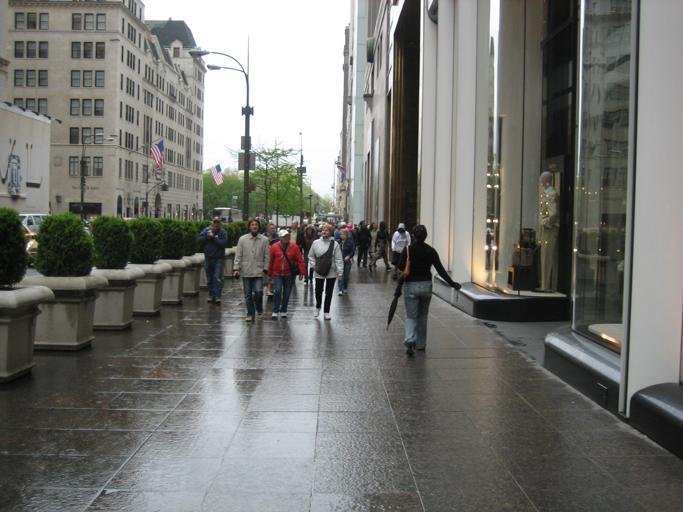}&
		\includegraphics[width=0.16\linewidth,height=0.8in,valign=m]{visual_evidence/input_0315.jpg} \\
		\addlinespace[3pt]
		\rotatebox[origin=c]{90}{\scriptsize{(b) GT}}
		\includegraphics[width=0.16\linewidth,height=0.8in,valign=m]{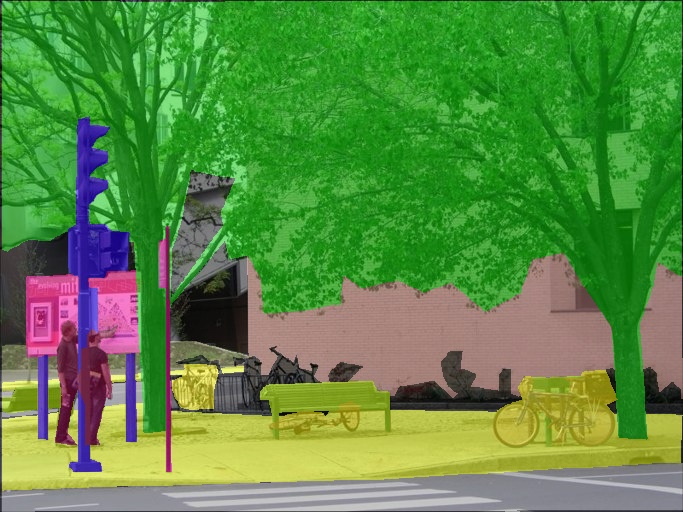}&
		\includegraphics[width=0.16\linewidth,height=0.8in,valign=m]{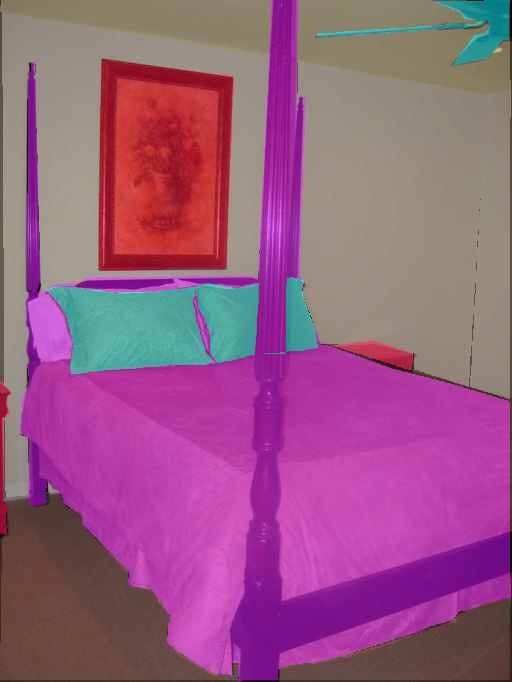}&
		\includegraphics[width=0.16\linewidth,height=0.8in,valign=m]{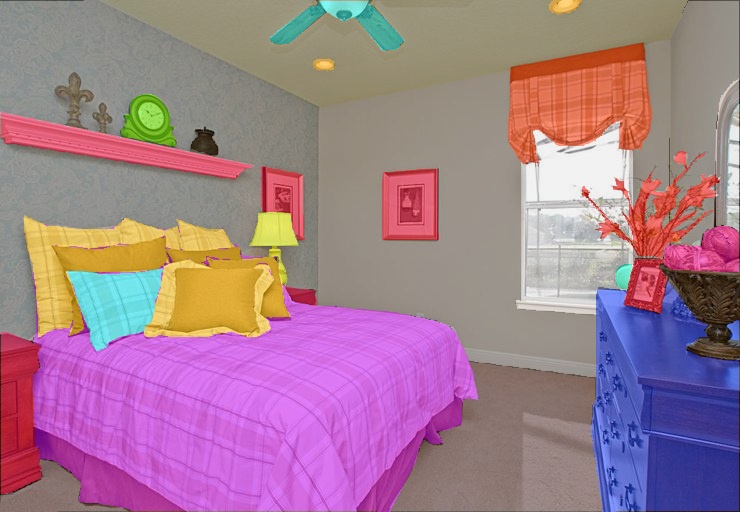}&
		\includegraphics[width=0.16\linewidth,height=0.8in,valign=m]{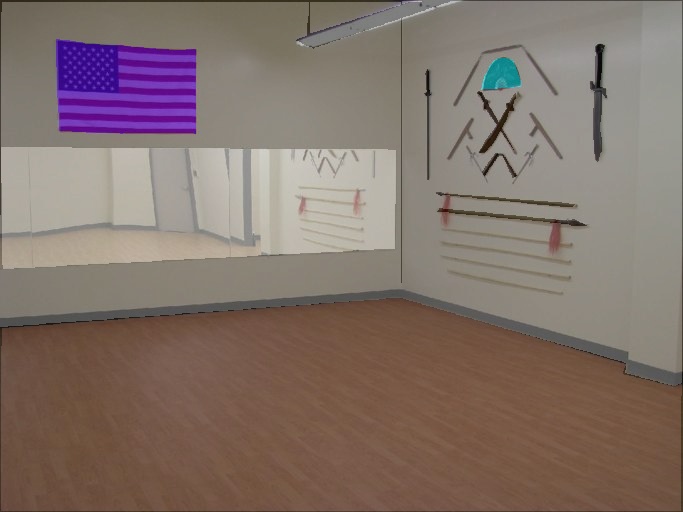}&
		\includegraphics[width=0.16\linewidth,height=0.8in,valign=m]{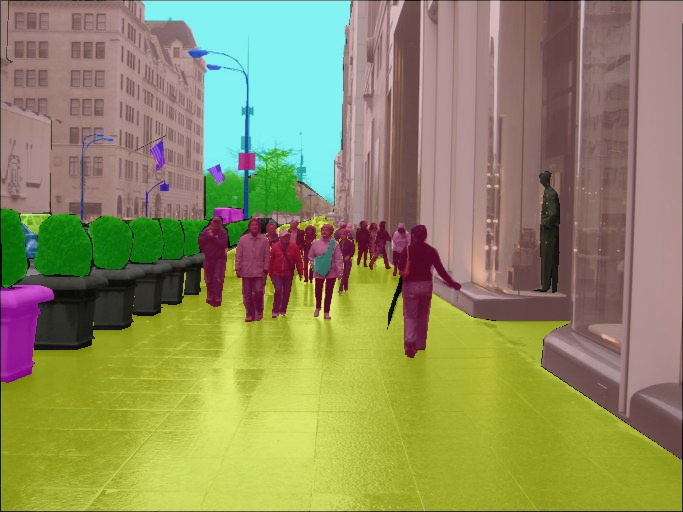}&
		\includegraphics[width=0.16\linewidth,height=0.8in,valign=m]{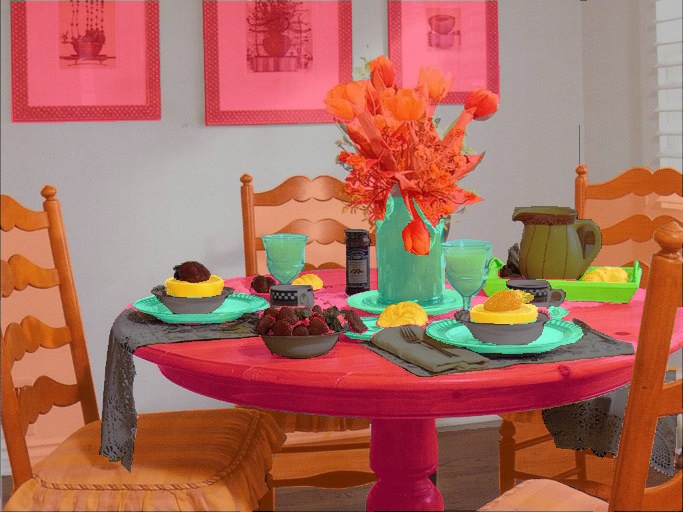} \\
		\addlinespace[3pt]
        \rotatebox[origin=c]{90}{\scriptsize{(c) Pixel Rebalance}}
	    \includegraphics[width=0.16\linewidth,height=0.8in,valign=m]{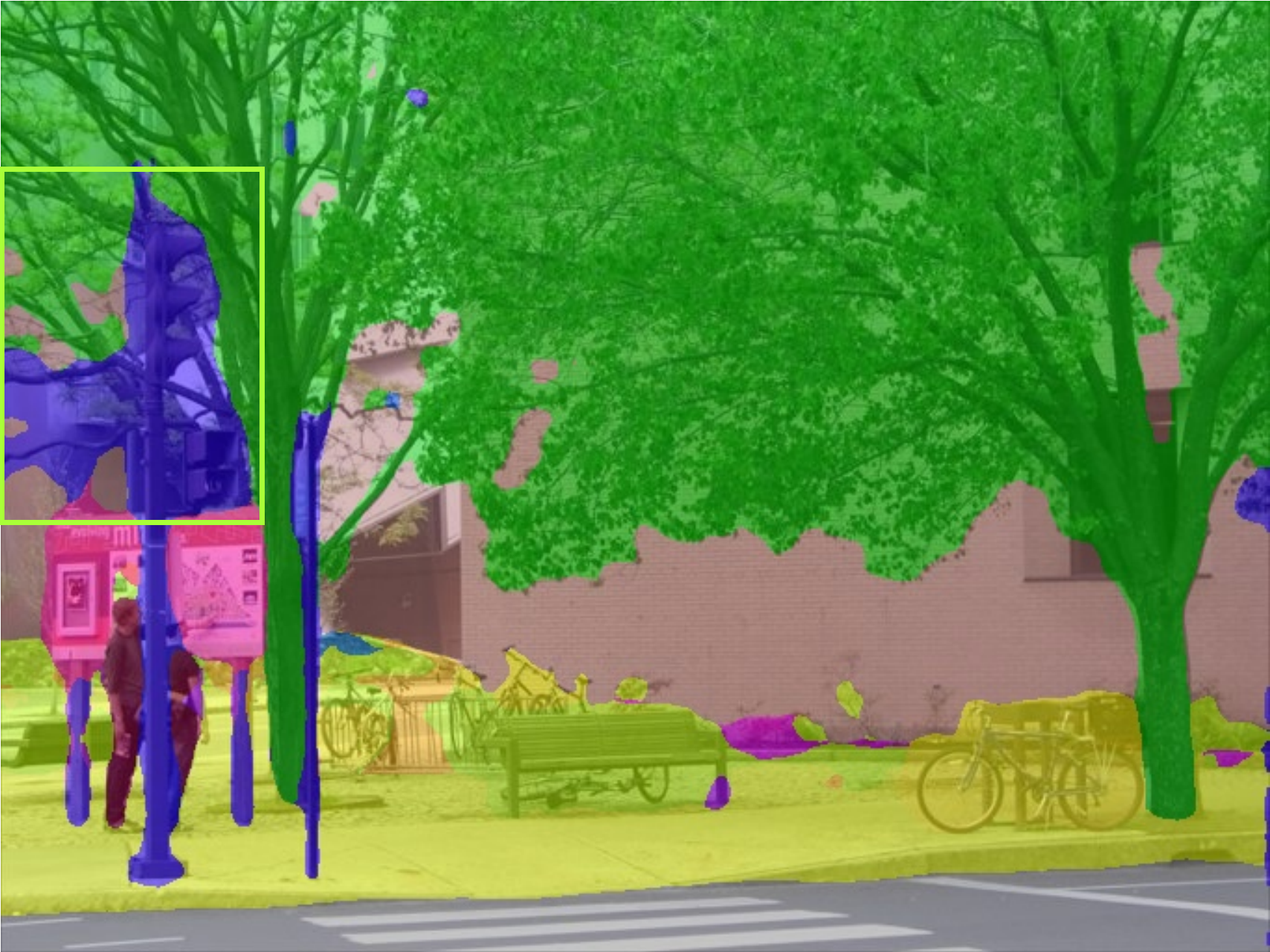}&
		\includegraphics[width=0.16\linewidth,height=0.8in,valign=m]{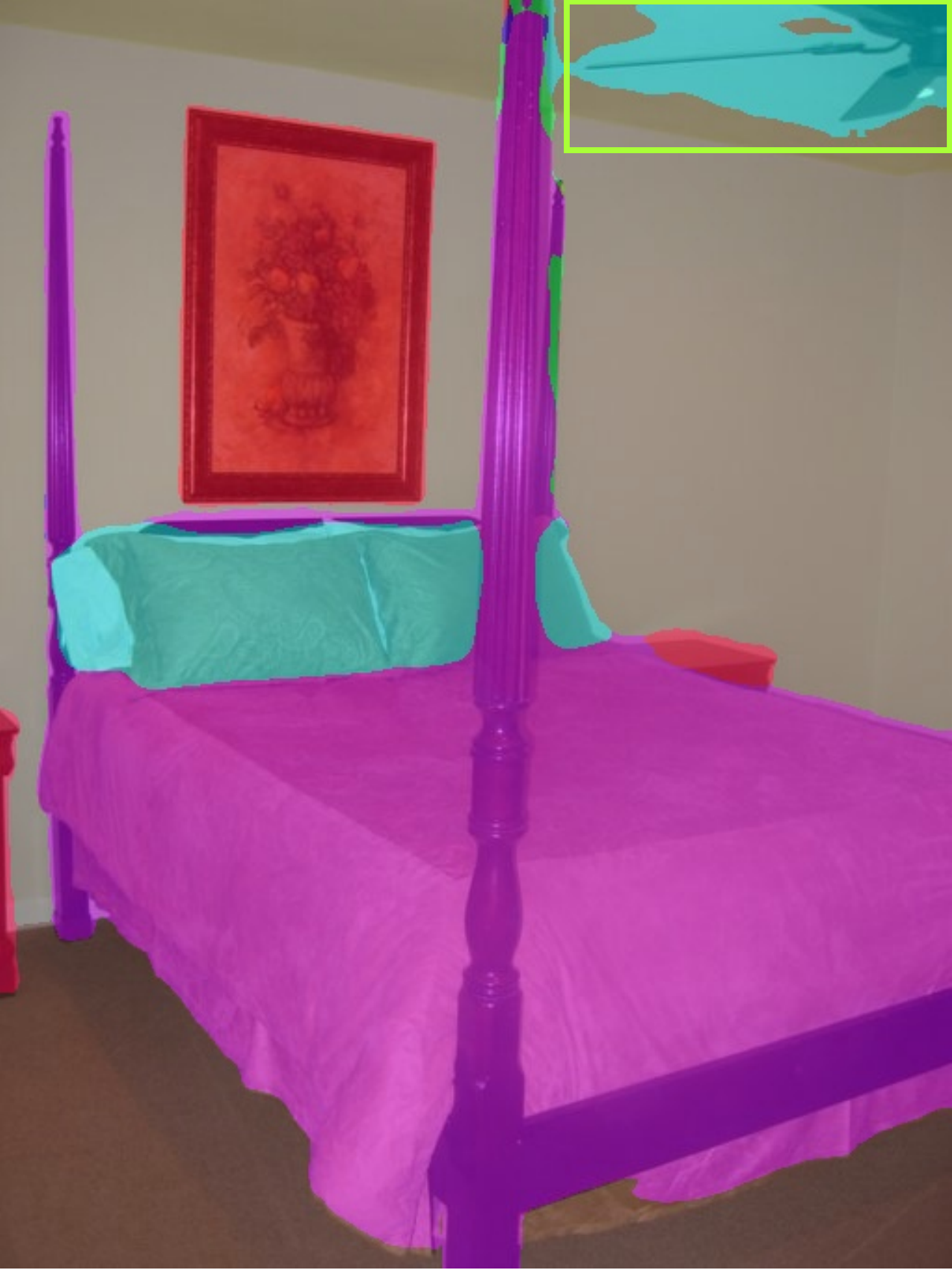}&
		\includegraphics[width=0.16\linewidth,height=0.8in,valign=m]{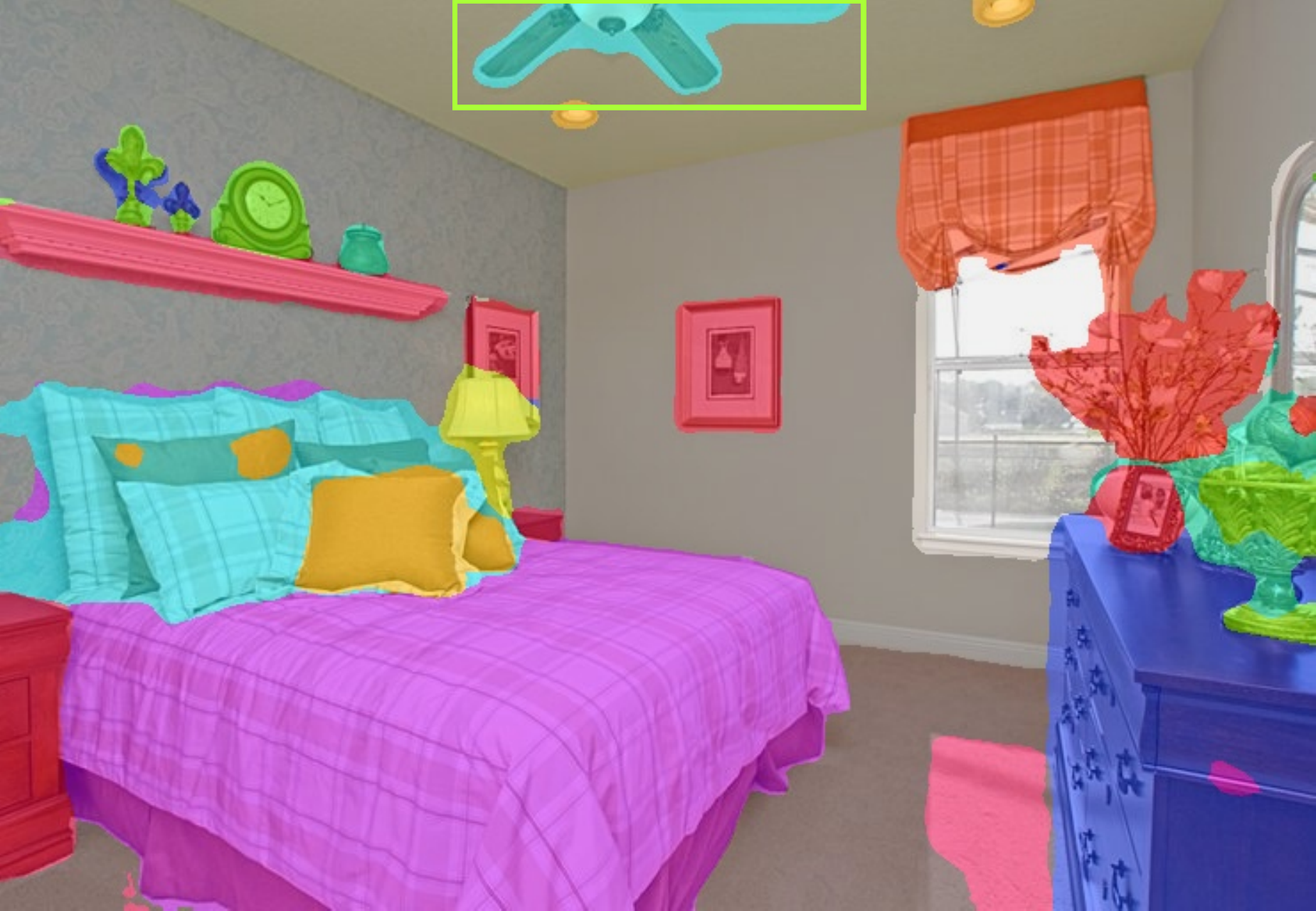}&
		\includegraphics[width=0.16\linewidth,height=0.8in,valign=m]{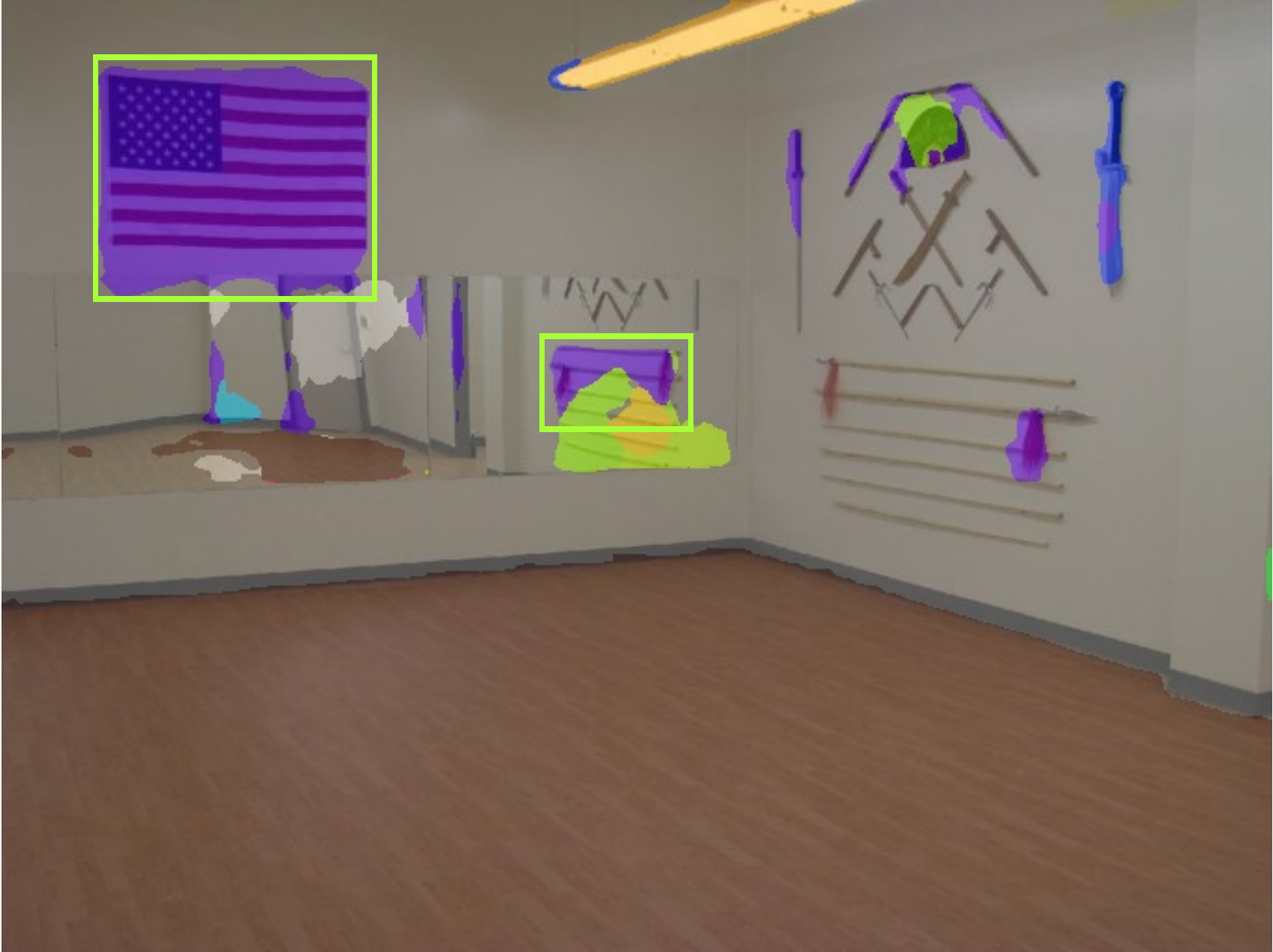}&
		\includegraphics[width=0.16\linewidth,height=0.8in,valign=m]{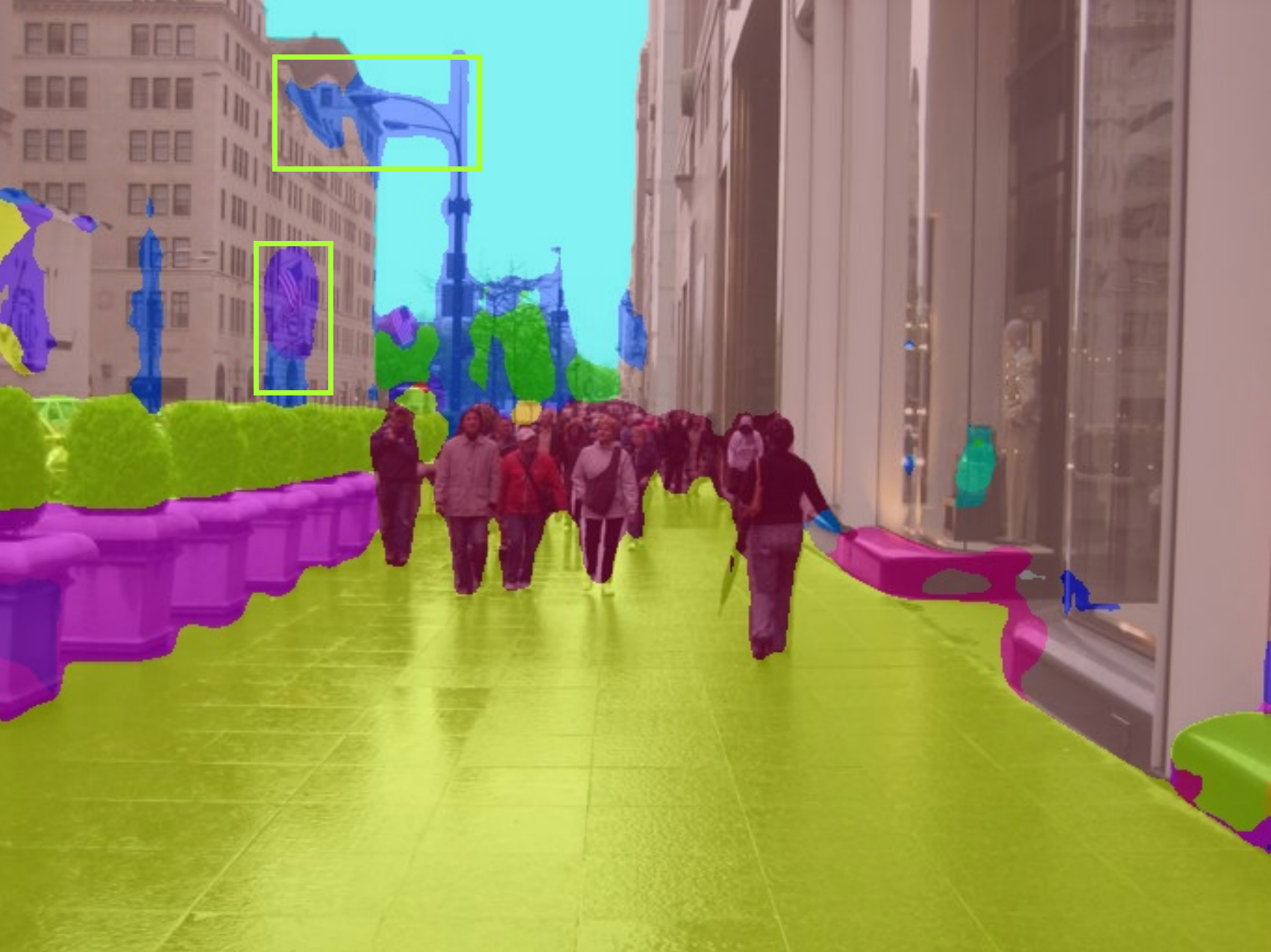}&
		\includegraphics[width=0.16\linewidth,height=0.8in,valign=m]{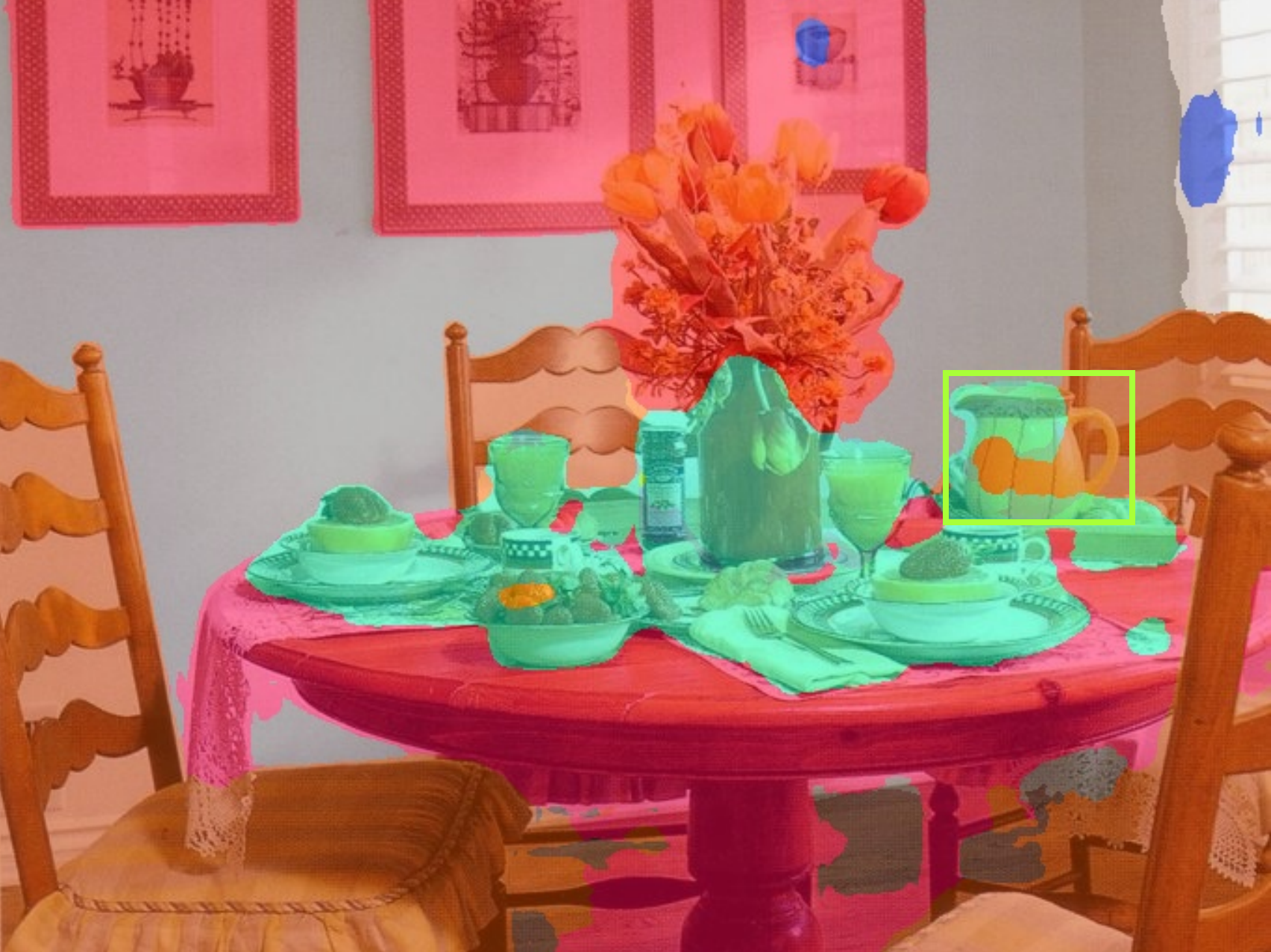} \\
	    \addlinespace[3pt]
	    \rotatebox[origin=c]{90}{\scriptsize{(d) Region Rebalance}}
		\includegraphics[width=0.16\linewidth,height=0.8in,valign=m]{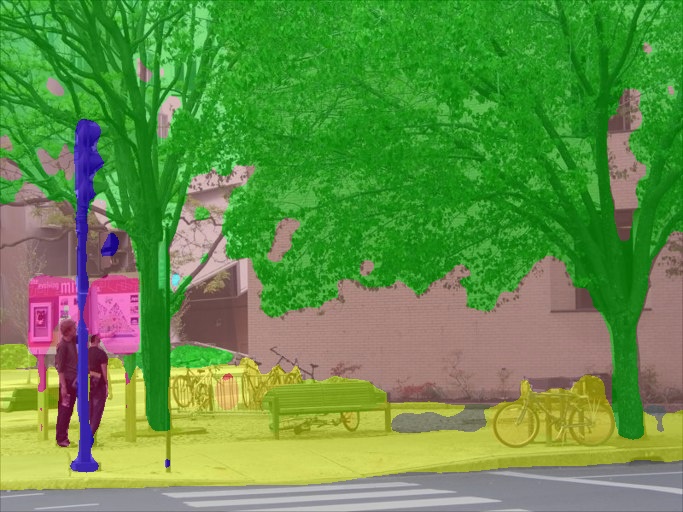}&
		\includegraphics[width=0.16\linewidth,height=0.8in,valign=m]{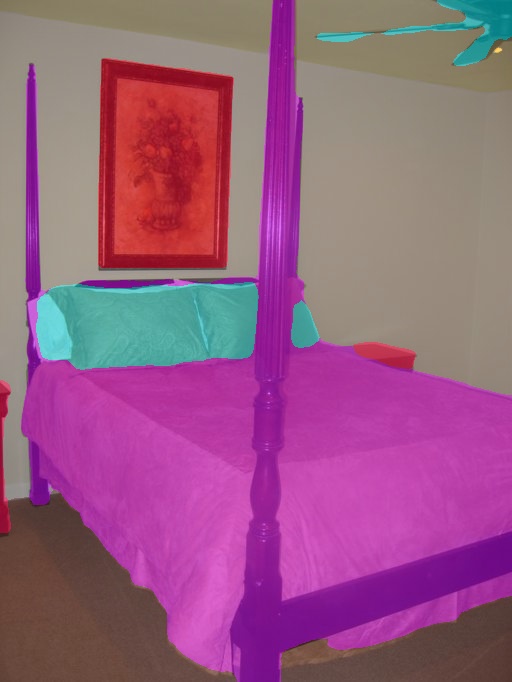}&
		\includegraphics[width=0.16\linewidth,height=0.8in,valign=m]{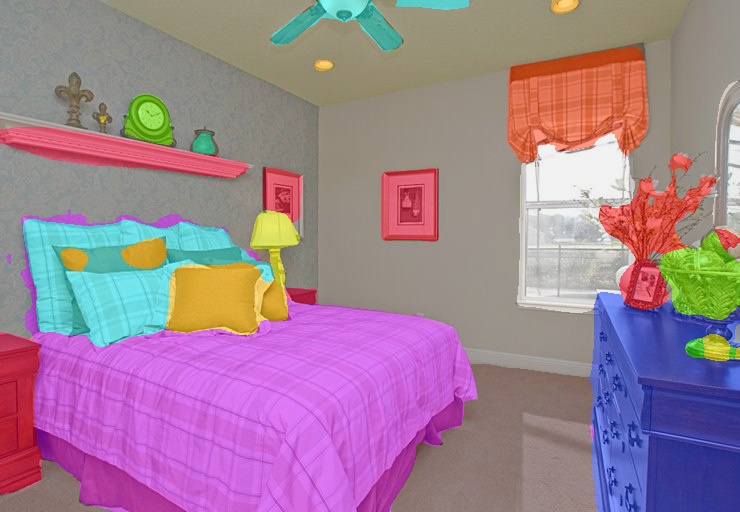}&
		\includegraphics[width=0.16\linewidth,height=0.8in,valign=m]{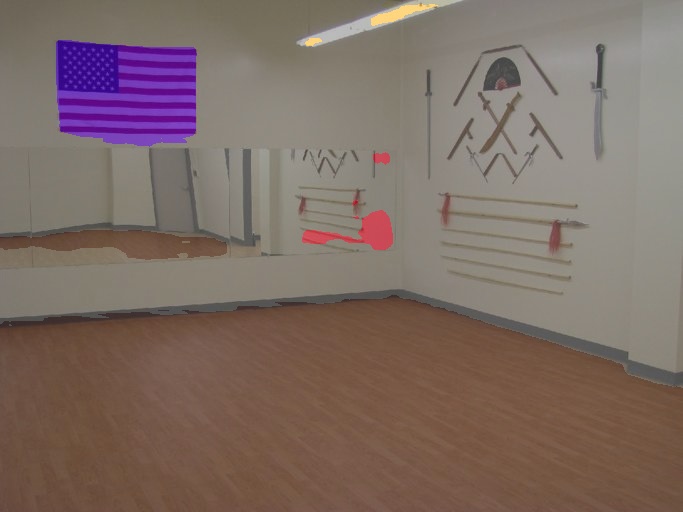}&
		\includegraphics[width=0.16\linewidth,height=0.8in,valign=m]{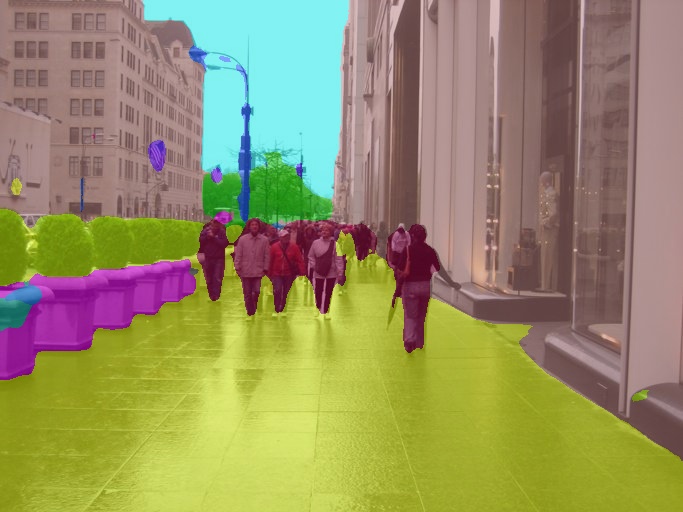}&
		\includegraphics[width=0.16\linewidth,height=0.8in,valign=m]{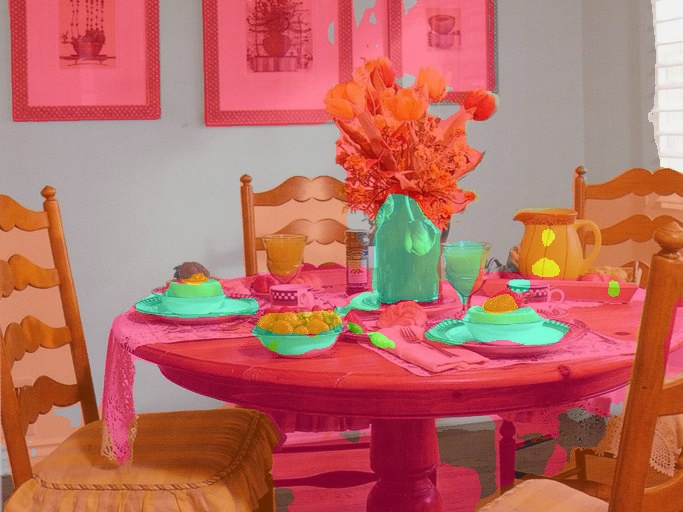} \\
	\end{tabular}}
	\caption{Visualization comparisons between pixel rebalance and region rebalance.}
	\label{fig:visual_evidence}
\end{figure*}

\noindent As demonstrated in Sec. \ref{sec:challenge1}, pixel rebalance improves $\mathrm{Acc}$ meanwhile increases $\mathrm{FP}$ significantly for low-frequency classes. As a result, $\mathrm{IoU}$ can not benefit from the pixel rebalance. In contrast, the proposed region rebalance is more aligned with the $\mathrm{IoU}$ metric. 
This phenomenon can be understood by considering the situation that two very different objects or stuff can still have similar local pixel features. Conducting pixel rebalance make it easy to let these similar pixels of other classes to be $\mathrm{FP}$. Taking it into consideration that objects/stuff are more distinguishable at the region level because of the global features, predicting other class objects/stuff as $\mathrm{FP}$ will be much harder. We give visual evidence to our intuitive reasoning in Figure \ref{fig:visual_evidence}.

\noindent As shown in highlighted regions with rectangle, pixels having similar texture with other object/stuff or around objects/stuff boundary are more likely to be $\mathrm{FP}$ with pixel rebalance. Region rebalance relieves this issue by using global region features.

\newpage
\section{Quantitative Visualization Comparisons on ADE$20$K and COCO-Stuff$164$K}
\begin{figure*}[h]
	\centering
	\resizebox{1.0\linewidth}{!}{
	\begin{tabular}{@{\hspace{0.5mm}}c@{\hspace{0.5mm}}c@{\hspace{0.5mm}}c@{\hspace{0.5mm}}c@{\hspace{0.5mm}}c@{\hspace{0.5mm}}c@{\hspace{0.0mm}}}
		\rotatebox[origin=c]{90}{\scriptsize{(a) Input}} 
	   \includegraphics[width=0.16\linewidth,height=0.8in,valign=m]{ade_vis/input_1145.jpg}&
		\includegraphics[width=0.16\linewidth,height=0.8in,valign=m]{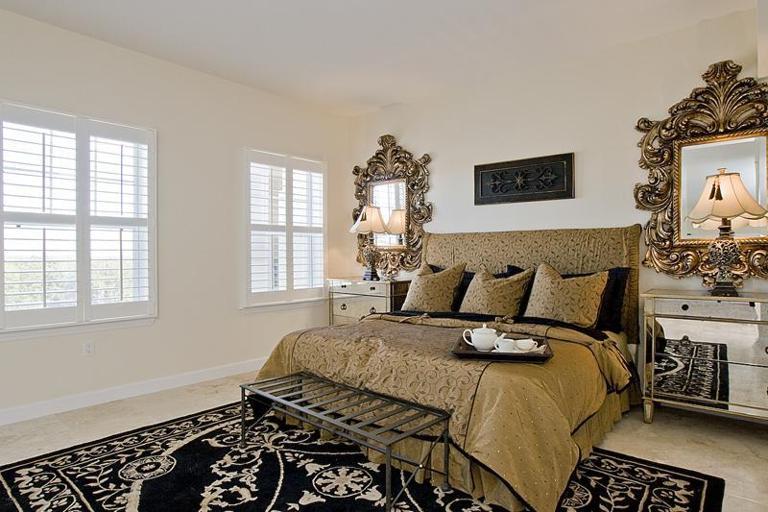}&
		\includegraphics[width=0.16\linewidth,height=0.8in,valign=m]{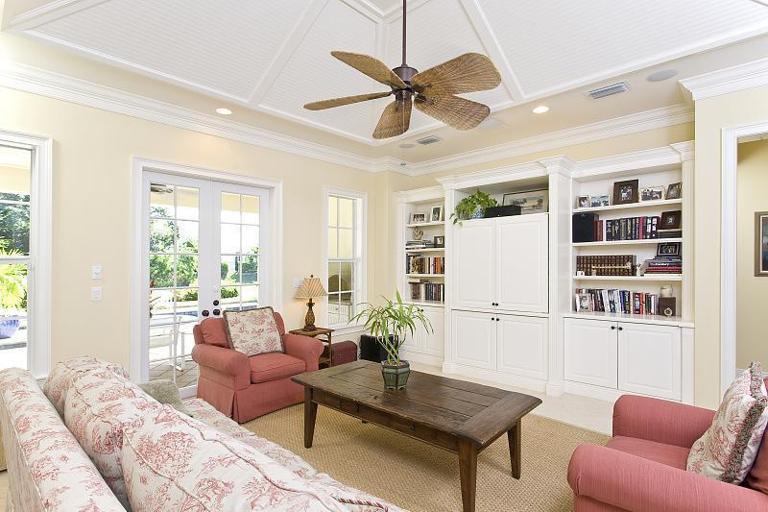}&
		\includegraphics[width=0.16\linewidth,height=0.8in,valign=m]{ade_vis/input_1671.jpg}&
		\includegraphics[width=0.16\linewidth,height=0.8in,valign=m]{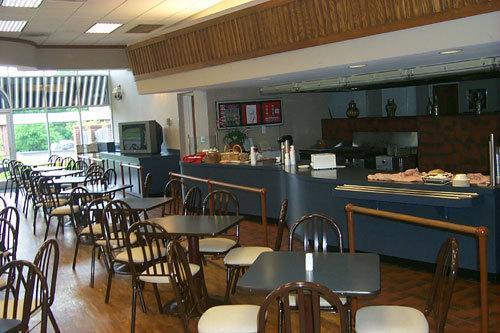}&
		\includegraphics[width=0.16\linewidth,height=0.8in,valign=m]{ade_vis/input_1698.jpg} \\
		\addlinespace[3pt]
		\rotatebox[origin=c]{90}{\scriptsize{(b) GT}}
		\includegraphics[width=0.16\linewidth,height=0.8in,valign=m]{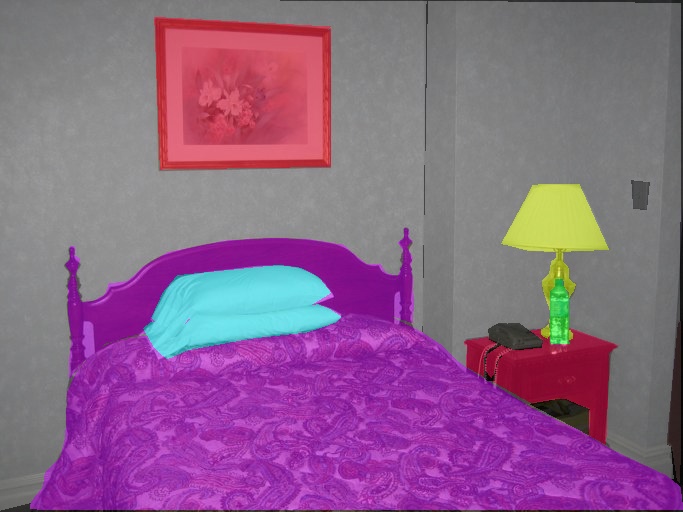}&
		\includegraphics[width=0.16\linewidth,height=0.8in,valign=m]{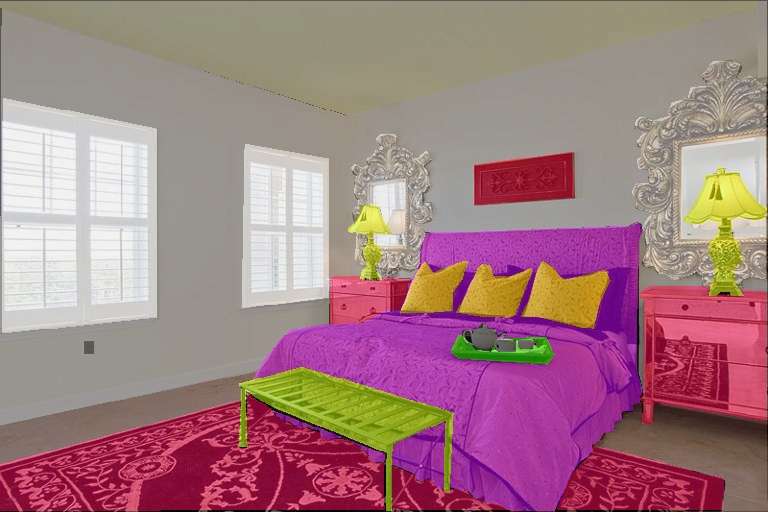}&
		\includegraphics[width=0.16\linewidth,height=0.8in,valign=m]{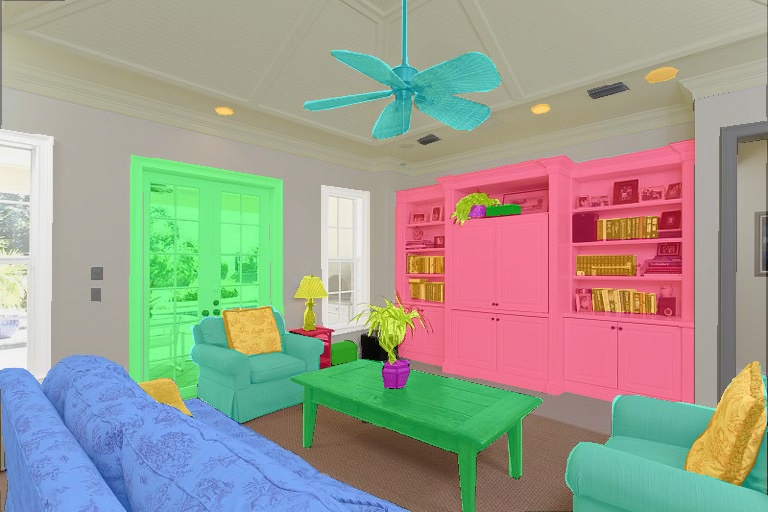}&
		\includegraphics[width=0.16\linewidth,height=0.8in,valign=m]{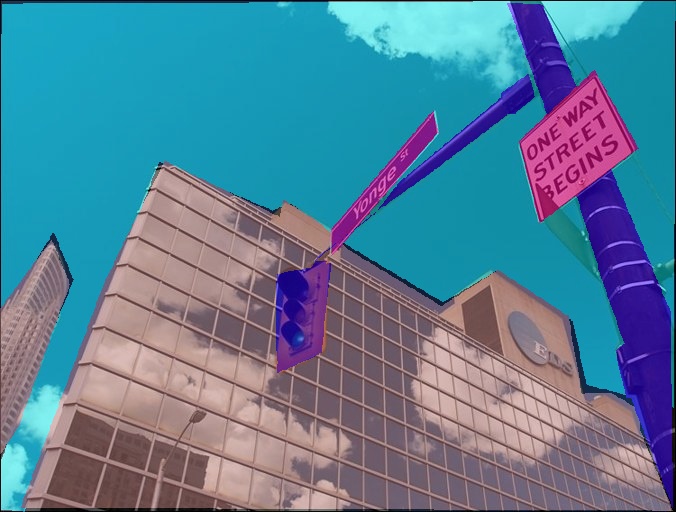}&
		\includegraphics[width=0.16\linewidth,height=0.8in,valign=m]{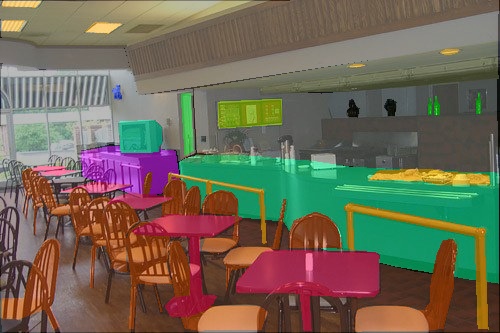}&
		\includegraphics[width=0.16\linewidth,height=0.8in,valign=m]{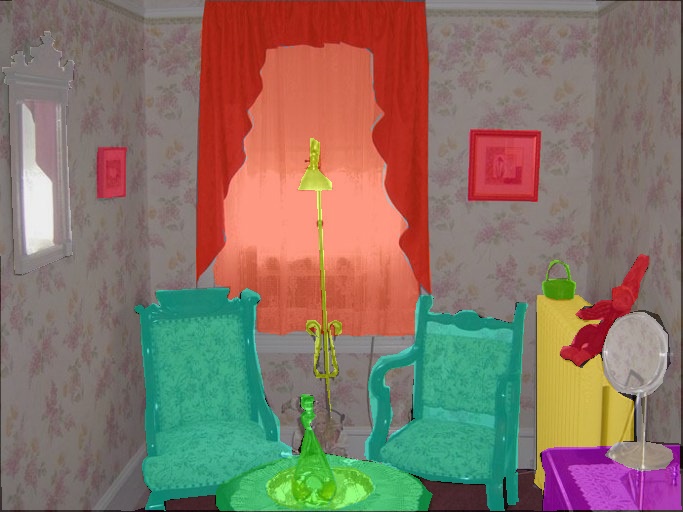} \\
		\addlinespace[3pt]
        \rotatebox[origin=c]{90}{\scriptsize{(c) Baseline}}
	    \includegraphics[width=0.16\linewidth,height=0.8in,valign=m]{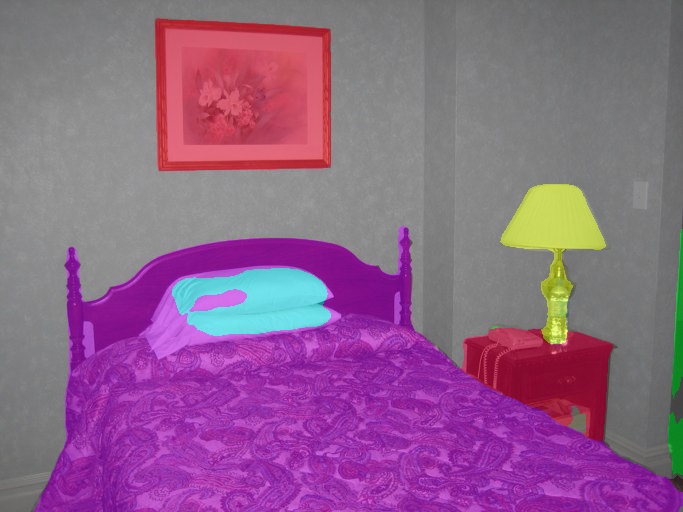}&
		\includegraphics[width=0.16\linewidth,height=0.8in,valign=m]{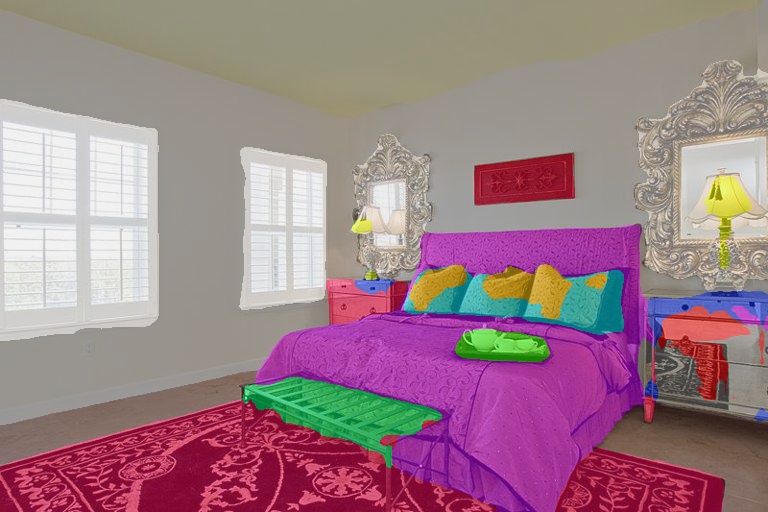}&
		\includegraphics[width=0.16\linewidth,height=0.8in,valign=m]{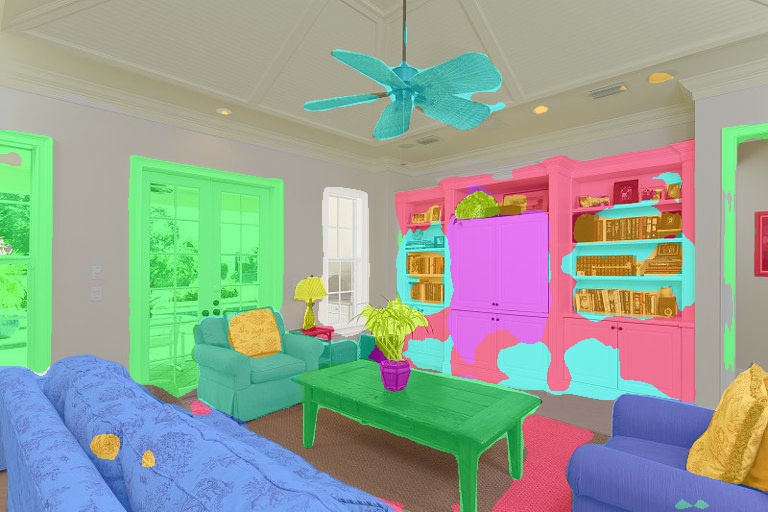}&
		\includegraphics[width=0.16\linewidth,height=0.8in,valign=m]{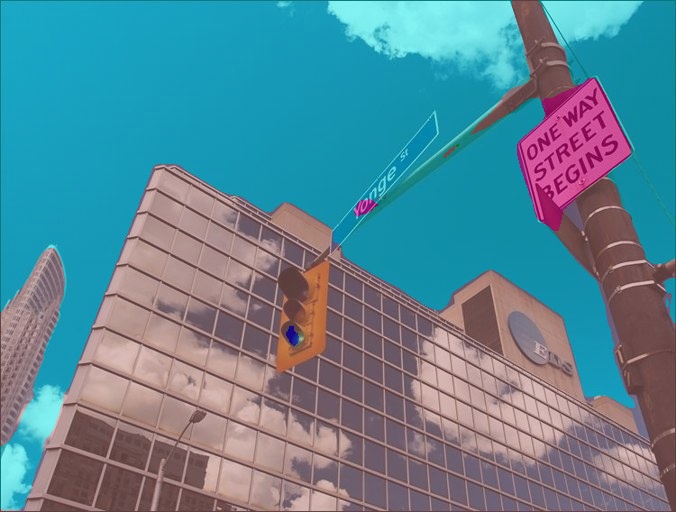}&
		\includegraphics[width=0.16\linewidth,height=0.8in,valign=m]{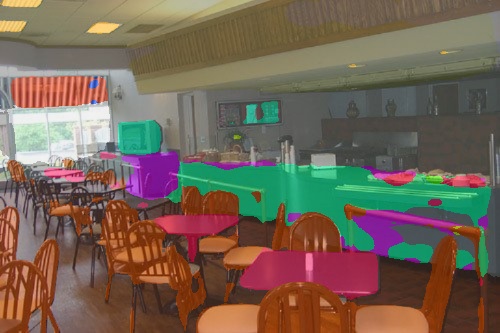}&
		\includegraphics[width=0.16\linewidth,height=0.8in,valign=m]{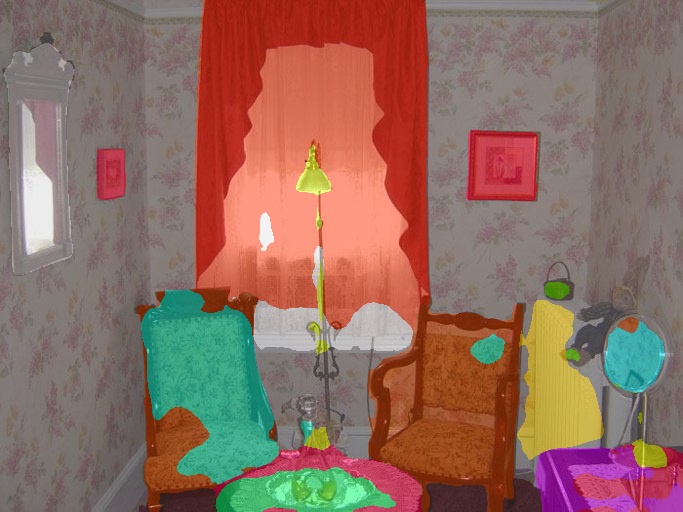} \\
	    \addlinespace[3pt]
	    \rotatebox[origin=c]{90}{\scriptsize{(d) Region Rebalance}}
		\includegraphics[width=0.16\linewidth,height=0.8in,valign=m]{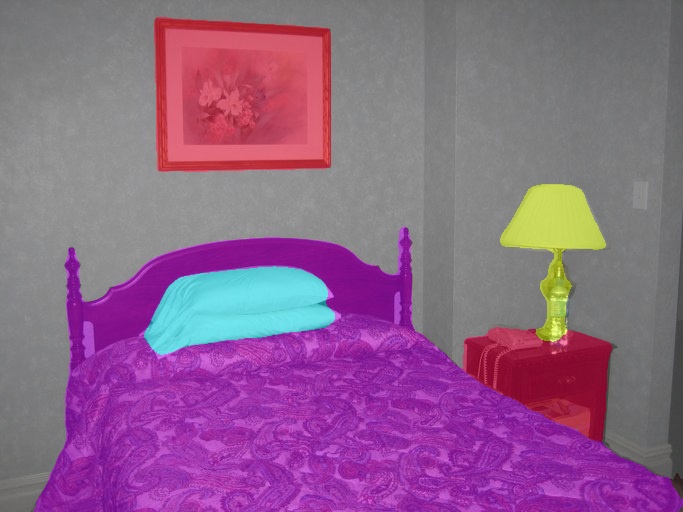}&
		\includegraphics[width=0.16\linewidth,height=0.8in,valign=m]{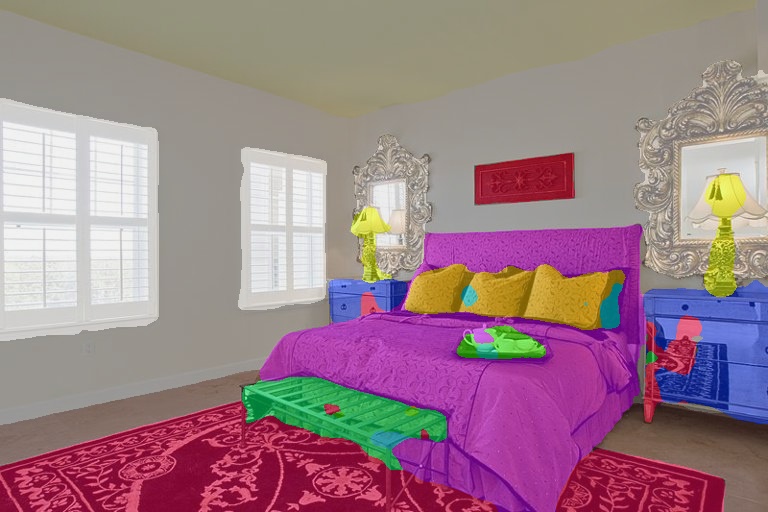}&
		\includegraphics[width=0.16\linewidth,height=0.8in,valign=m]{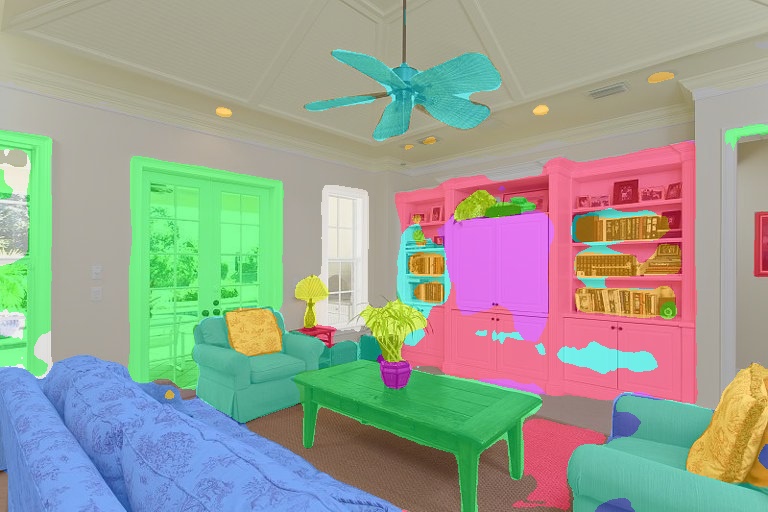}&
		\includegraphics[width=0.16\linewidth,height=0.8in,valign=m]{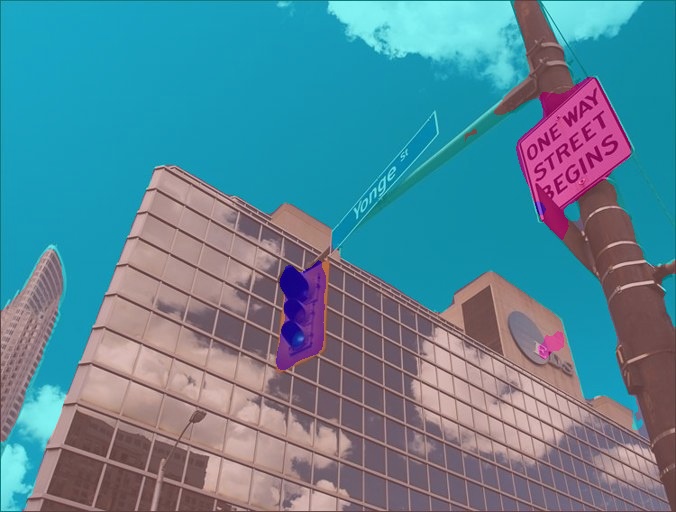}&
		\includegraphics[width=0.16\linewidth,height=0.8in,valign=m]{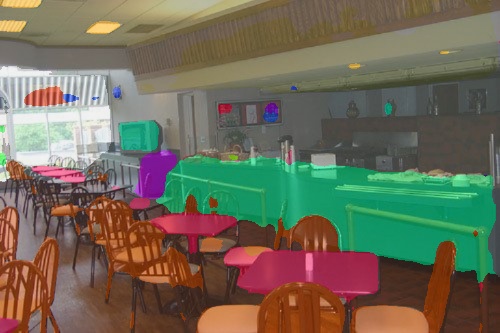}&
		\includegraphics[width=0.16\linewidth,height=0.8in,valign=m]{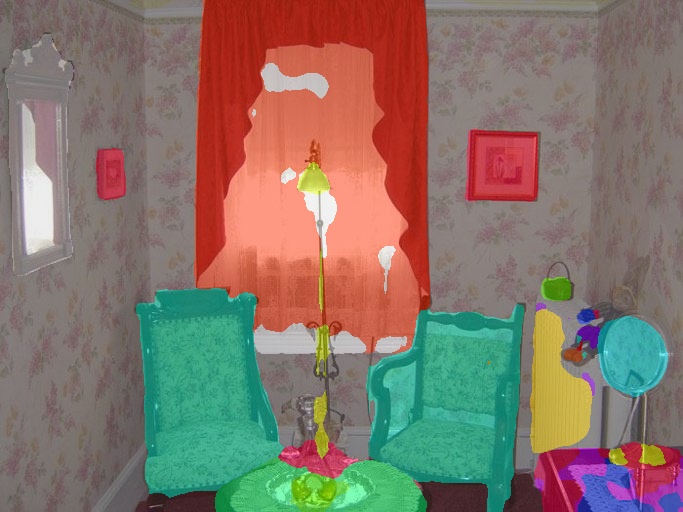} \\
	\end{tabular}}
	\caption{Visualization comparisons between baseline and region rebalance on ADE$20$K.}
	\label{fig:visual_comparison_ade20k_v1}
\end{figure*}

\begin{figure*}[h]
	\centering
	\resizebox{1.0\linewidth}{!}{
	\begin{tabular}{@{\hspace{0.5mm}}c@{\hspace{0.5mm}}c@{\hspace{0.5mm}}c@{\hspace{0.5mm}}c@{\hspace{0.5mm}}c@{\hspace{0.5mm}}c@{\hspace{0.0mm}}}
		\rotatebox[origin=c]{90}{\scriptsize{(a) Input}} 
	    \includegraphics[width=0.16\linewidth,height=0.8in,valign=m]{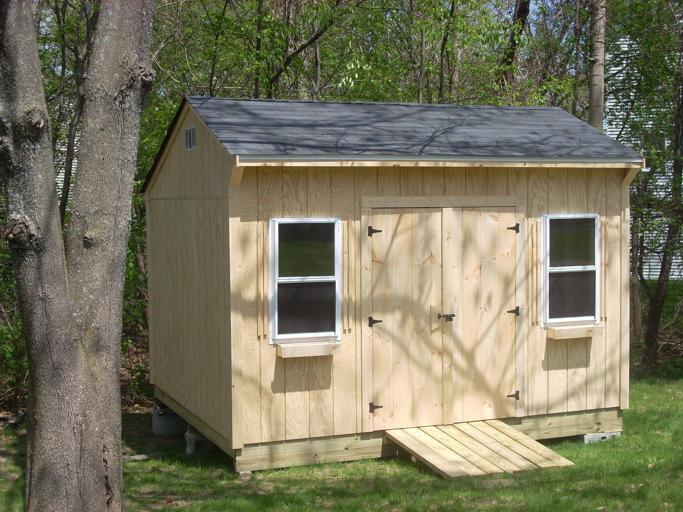}&
		\includegraphics[width=0.16\linewidth,height=0.8in,valign=m]{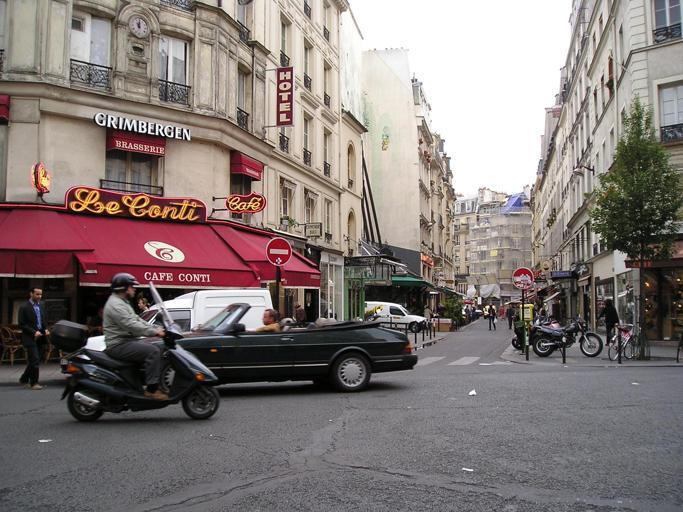}&
		\includegraphics[width=0.16\linewidth,height=0.8in,valign=m]{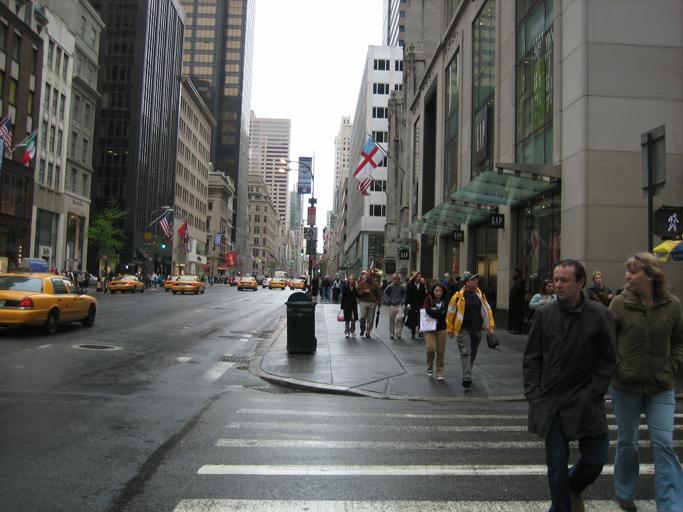}&
		\includegraphics[width=0.16\linewidth,height=0.8in,valign=m]{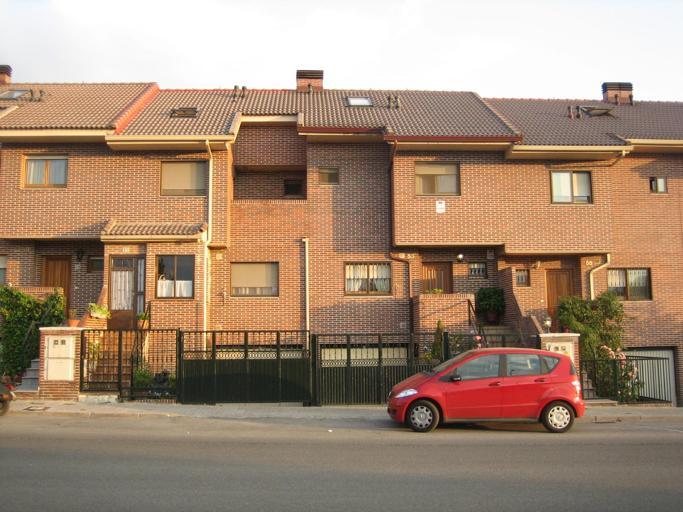}&
		\includegraphics[width=0.16\linewidth,height=0.8in,valign=m]{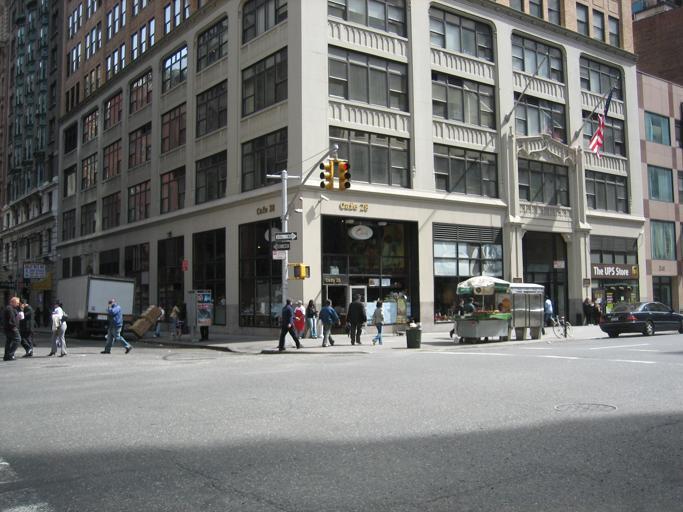}&
		\includegraphics[width=0.16\linewidth,height=0.8in,valign=m]{ade_vis/input_1827.jpg} \\
		\addlinespace[3pt]
		\rotatebox[origin=c]{90}{\scriptsize{(b) GT}}
		\includegraphics[width=0.16\linewidth,height=0.8in,valign=m]{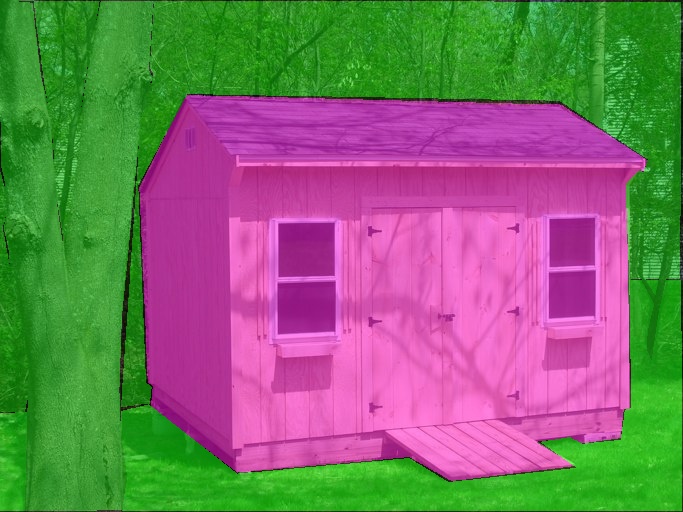}&
		\includegraphics[width=0.16\linewidth,height=0.8in,valign=m]{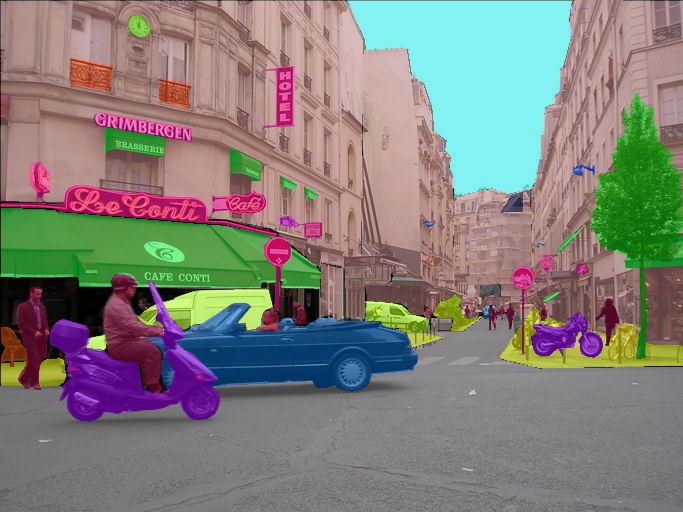}&
		\includegraphics[width=0.16\linewidth,height=0.8in,valign=m]{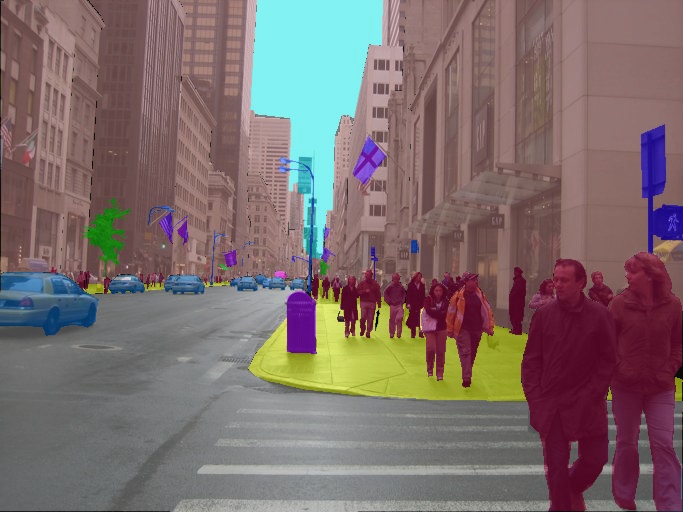}&
		\includegraphics[width=0.16\linewidth,height=0.8in,valign=m]{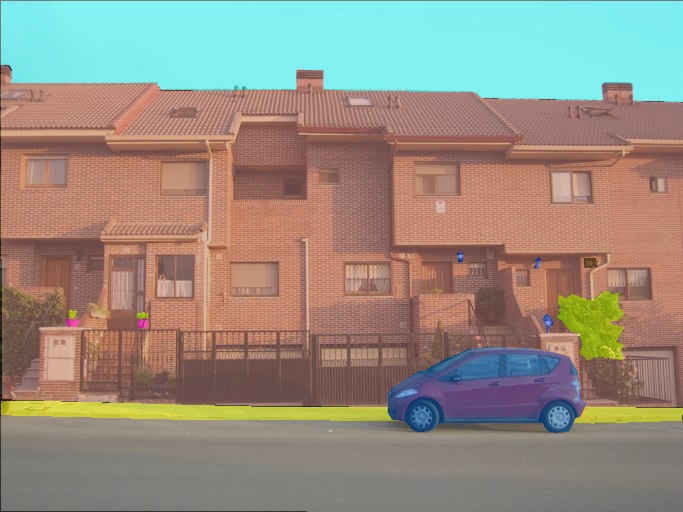}&
		\includegraphics[width=0.16\linewidth,height=0.8in,valign=m]{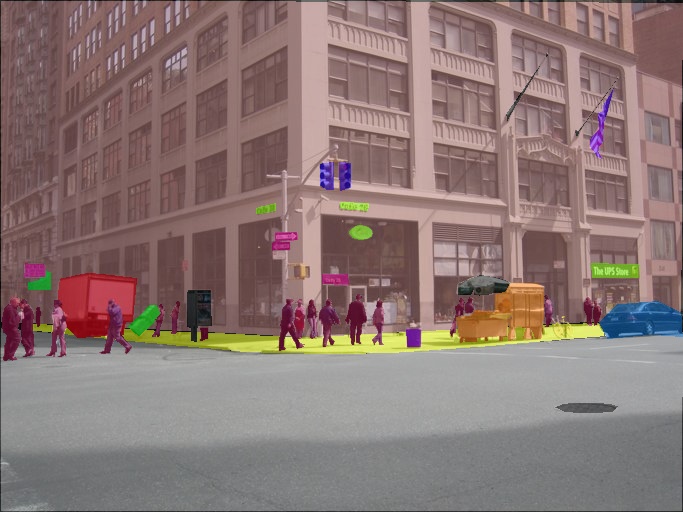}&
		\includegraphics[width=0.16\linewidth,height=0.8in,valign=m]{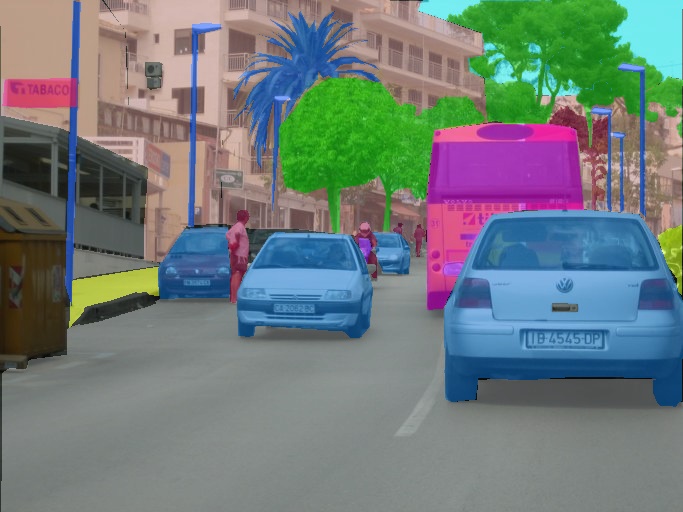} \\
		\addlinespace[3pt]
        \rotatebox[origin=c]{90}{\scriptsize{(c) Baseline}}
	    \includegraphics[width=0.16\linewidth,height=0.8in,valign=m]{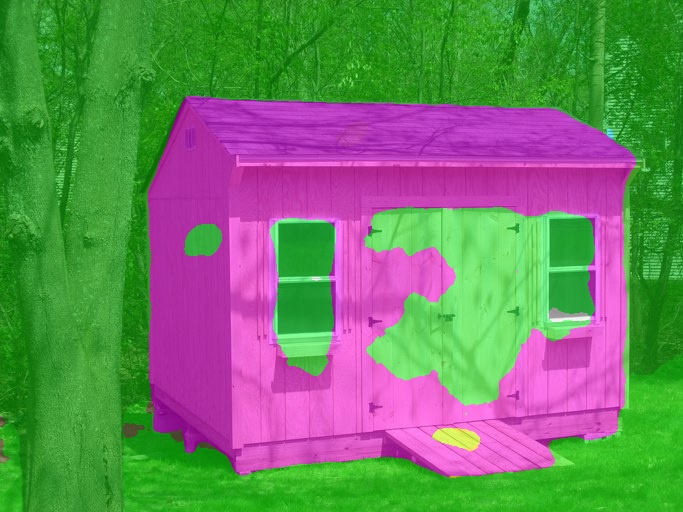}&
		\includegraphics[width=0.16\linewidth,height=0.8in,valign=m]{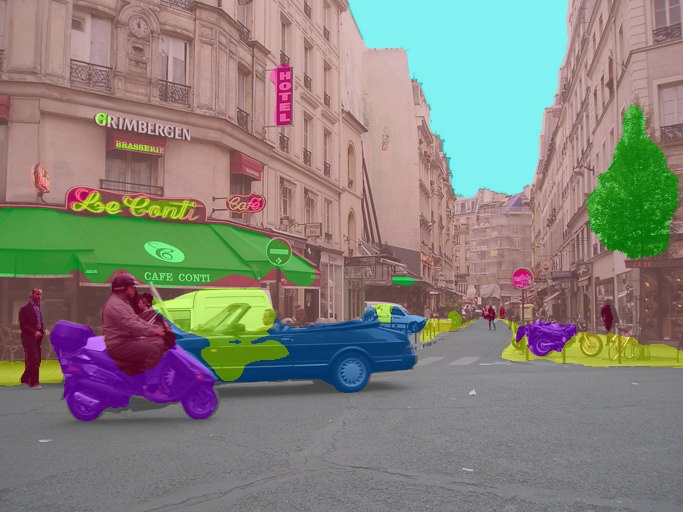}&
		\includegraphics[width=0.16\linewidth,height=0.8in,valign=m]{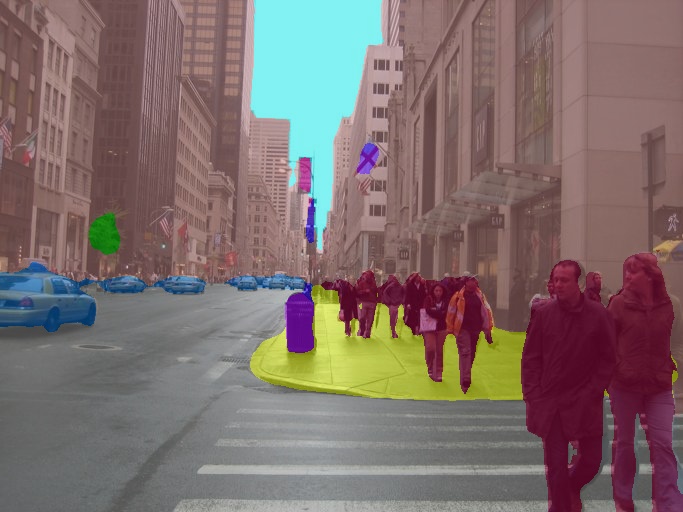}&
		\includegraphics[width=0.16\linewidth,height=0.8in,valign=m]{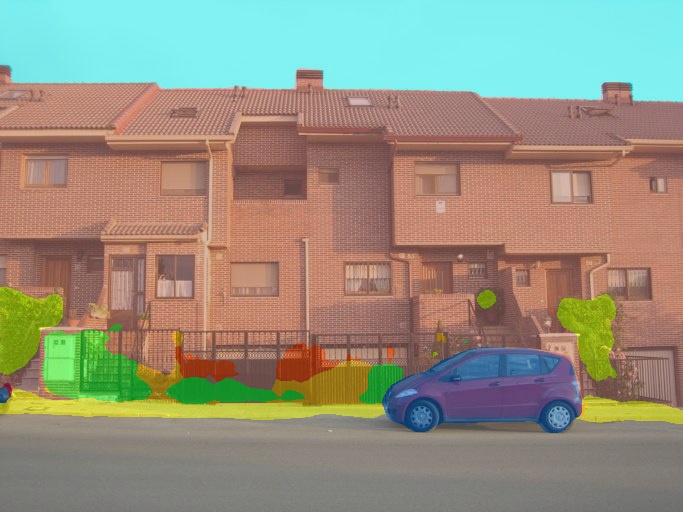}&
		\includegraphics[width=0.16\linewidth,height=0.8in,valign=m]{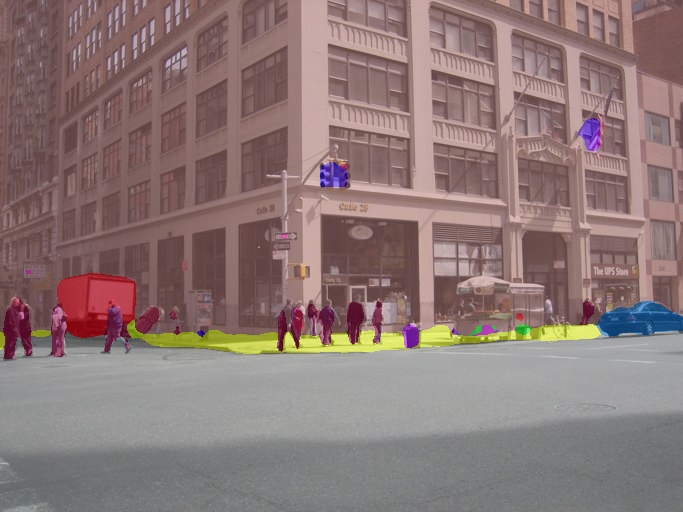}&
		\includegraphics[width=0.16\linewidth,height=0.8in,valign=m]{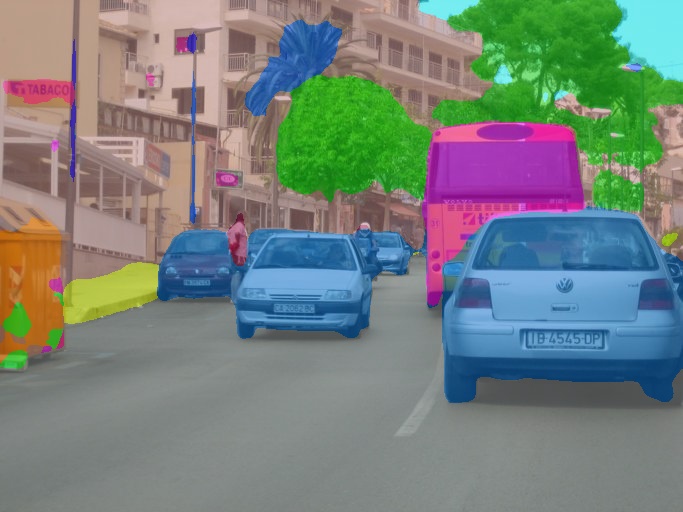} \\
	    \addlinespace[3pt]
	    \rotatebox[origin=c]{90}{\scriptsize{(d) Region Rebalance}}
		\includegraphics[width=0.16\linewidth,height=0.8in,valign=m]{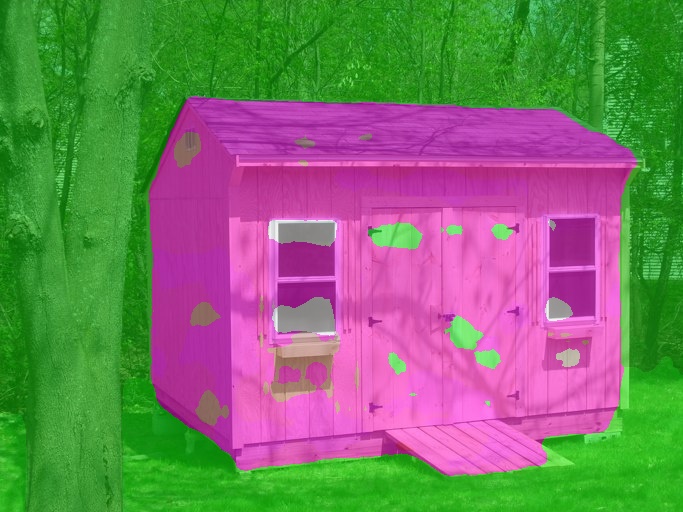}&
		\includegraphics[width=0.16\linewidth,height=0.8in,valign=m]{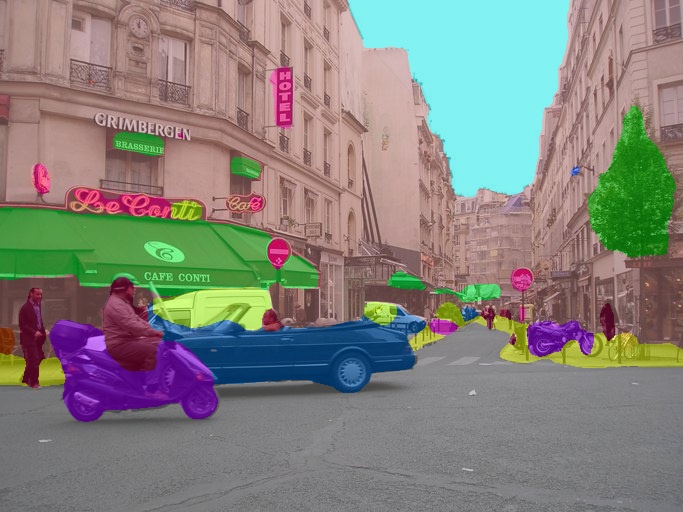}&
		\includegraphics[width=0.16\linewidth,height=0.8in,valign=m]{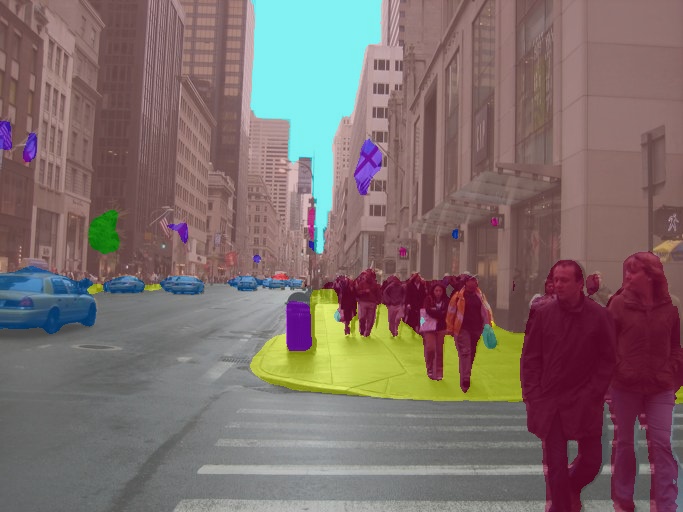}&
		\includegraphics[width=0.16\linewidth,height=0.8in,valign=m]{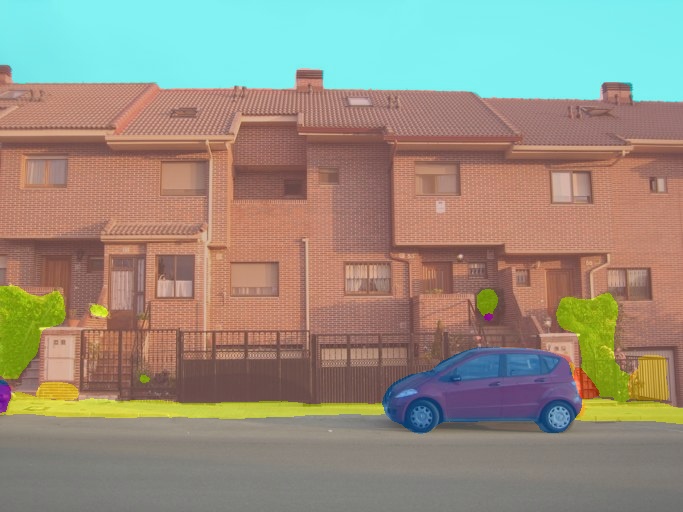}&
		\includegraphics[width=0.16\linewidth,height=0.8in,valign=m]{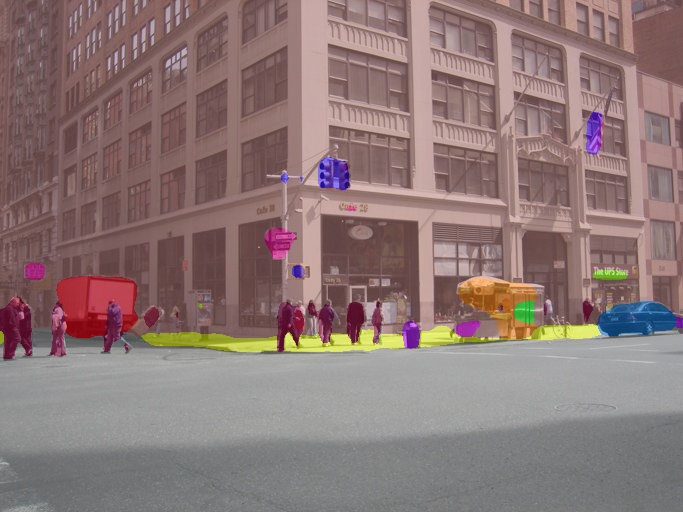}&
		\includegraphics[width=0.16\linewidth,height=0.8in,valign=m]{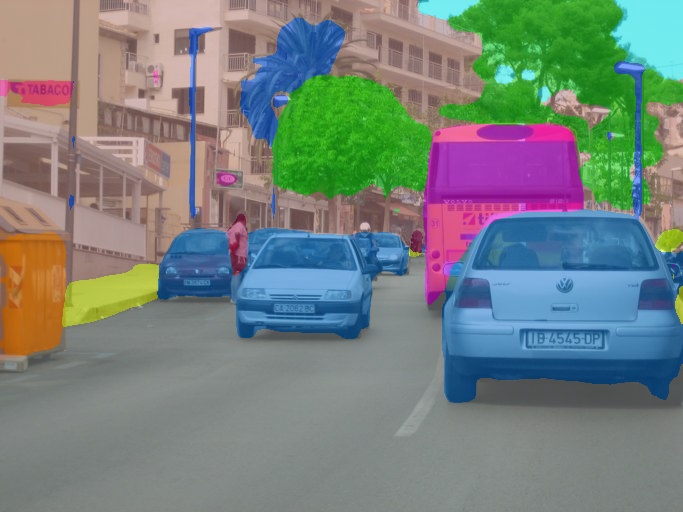} \\
	\end{tabular}}
	\caption{Visualization comparisons between baseline and region rebalance on ADE$20$K.}
	\label{fig:visual_comparison_ade20k_v2}
\end{figure*}

\begin{figure*}[h]
	\centering
	\resizebox{1.0\linewidth}{!}{
	\begin{tabular}{@{\hspace{0.5mm}}c@{\hspace{0.5mm}}c@{\hspace{0.5mm}}c@{\hspace{0.5mm}}c@{\hspace{0.5mm}}c@{\hspace{0.5mm}}c@{\hspace{0.5mm}}c@{\hspace{0.5mm}}c@{\hspace{0.0mm}}}
		\rotatebox[origin=c]{90}{\scriptsize{(a) Input}} 
        \includegraphics[width=0.16\linewidth,height=0.8in,valign=m]{coco_vis/input_570756.jpg}&
	    \includegraphics[width=0.16\linewidth,height=0.8in,valign=m]{coco_vis/input_446522.jpg}&
		\includegraphics[width=0.16\linewidth,height=0.8in,valign=m]{coco_vis/input_522713.jpg}&
		\includegraphics[width=0.16\linewidth,height=0.8in,valign=m]{coco_vis/input_562581.jpg}&
		\includegraphics[width=0.16\linewidth,height=0.8in,valign=m]{coco_vis/input_568290.jpg}&
		\includegraphics[width=0.16\linewidth,height=0.8in,valign=m]{coco_vis/input_569972.jpg} \\
		\addlinespace[3pt]
		\rotatebox[origin=c]{90}{\scriptsize{(b) GT}}
	    \includegraphics[width=0.16\linewidth,height=0.8in,valign=m]{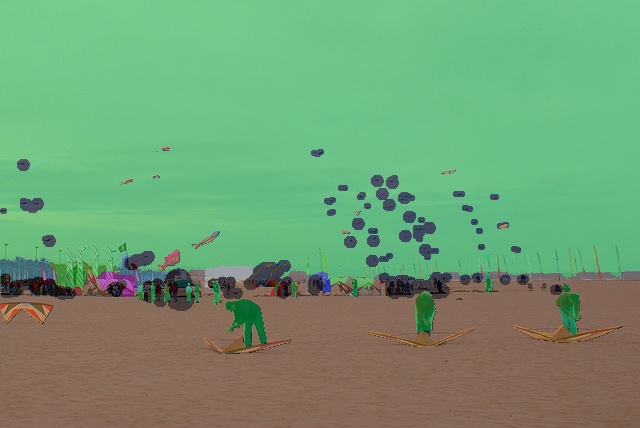}&
		\includegraphics[width=0.16\linewidth,height=0.8in,valign=m]{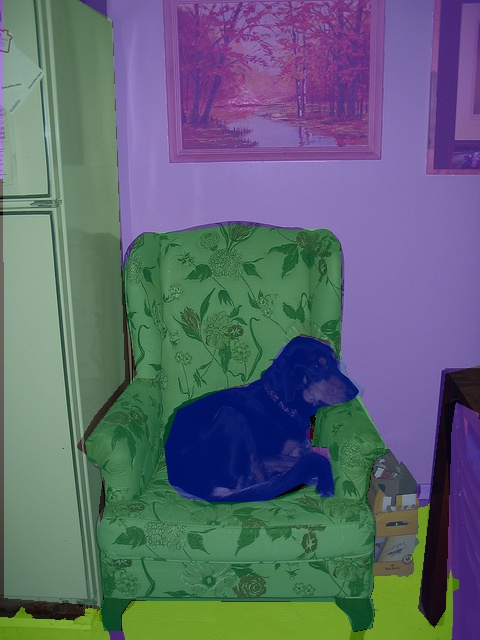}&
		\includegraphics[width=0.16\linewidth,height=0.8in,valign=m]{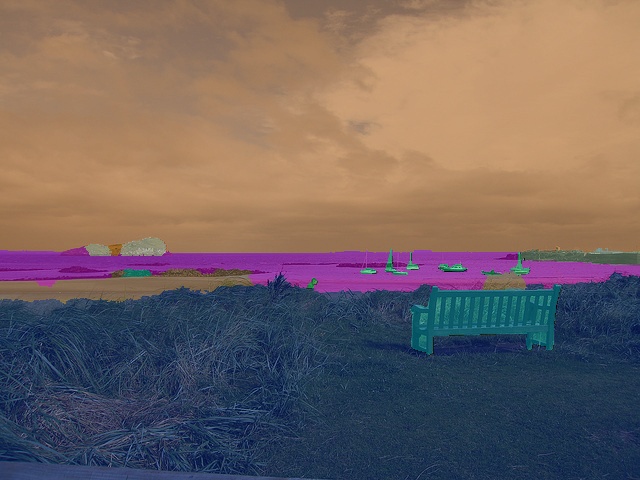}&
		\includegraphics[width=0.16\linewidth,height=0.8in,valign=m]{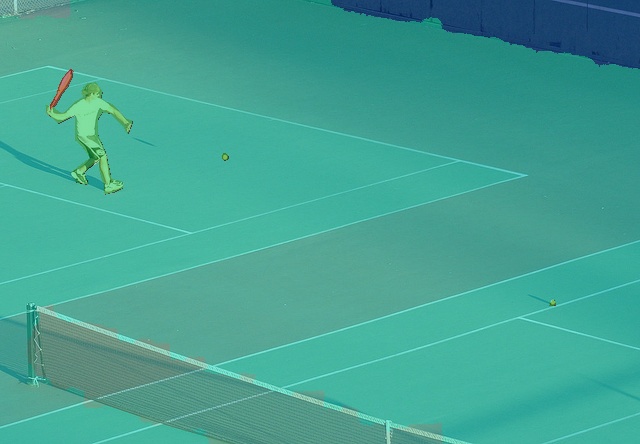}&
		\includegraphics[width=0.16\linewidth,height=0.8in,valign=m]{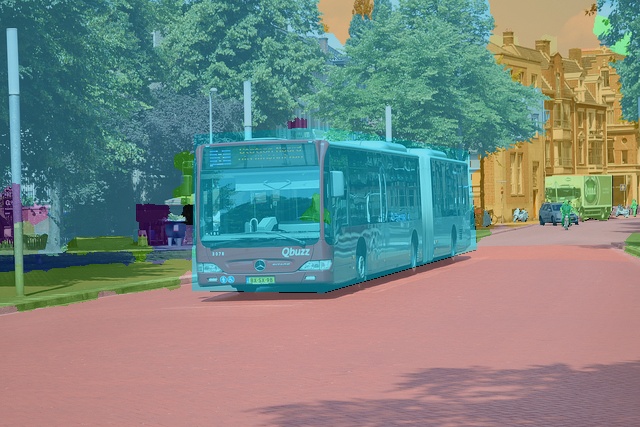}&
		\includegraphics[width=0.16\linewidth,height=0.8in,valign=m]{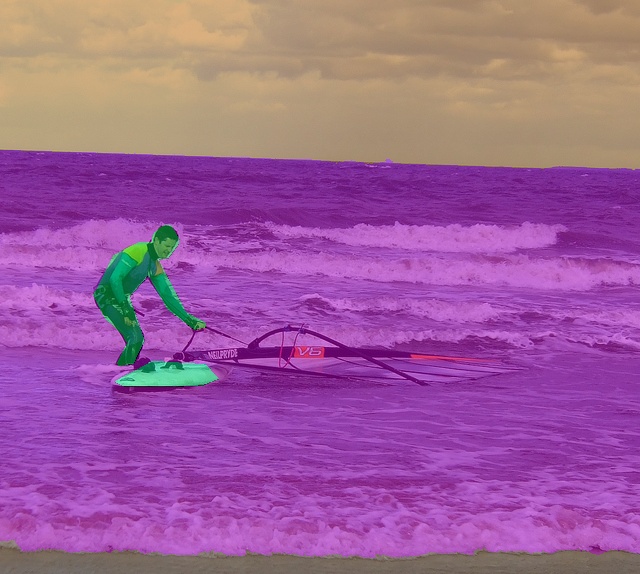} \\
		\addlinespace[3pt]
        \rotatebox[origin=c]{90}{\scriptsize{(c) Baseline}}
        \includegraphics[width=0.16\linewidth,height=0.8in,valign=m]{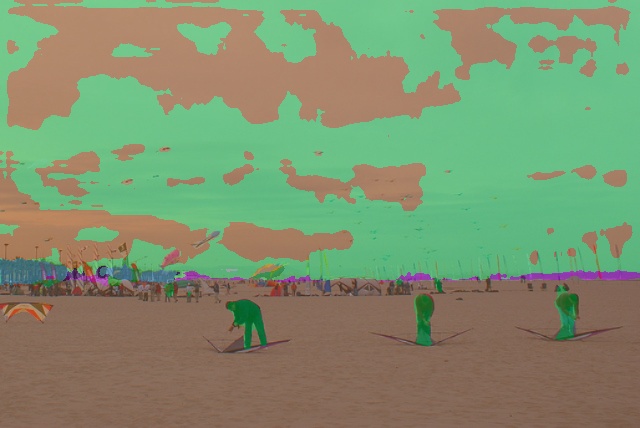}&
	    \includegraphics[width=0.16\linewidth,height=0.8in,valign=m]{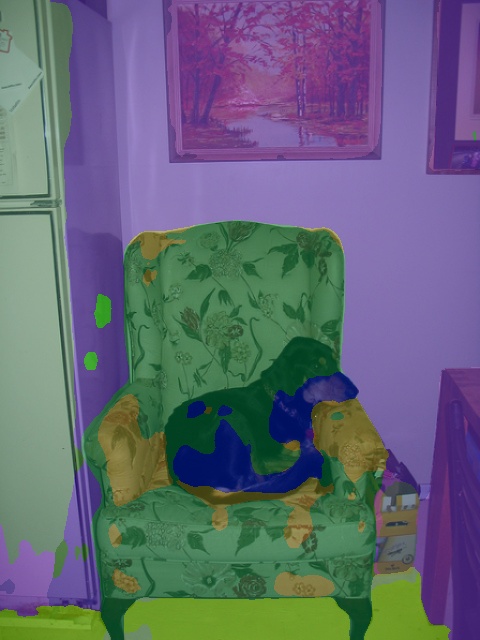}&
		\includegraphics[width=0.16\linewidth,height=0.8in,valign=m]{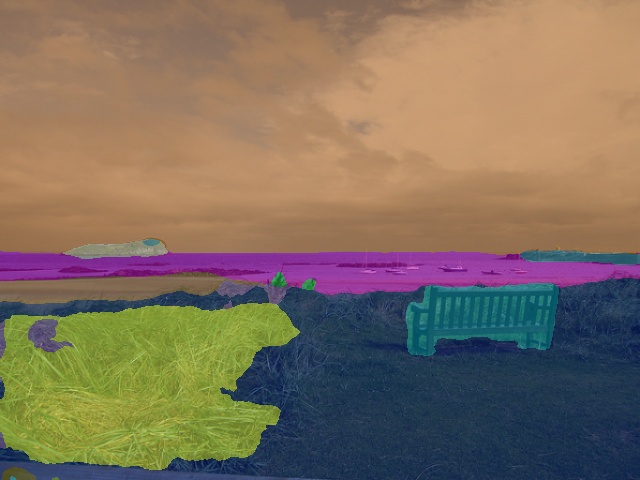}&
		\includegraphics[width=0.16\linewidth,height=0.8in,valign=m]{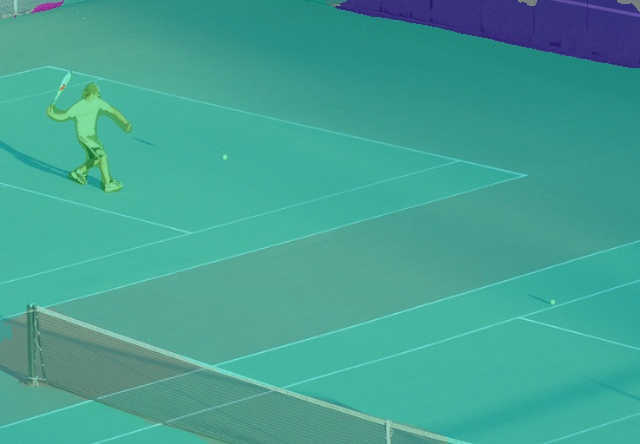}&
		\includegraphics[width=0.16\linewidth,height=0.8in,valign=m]{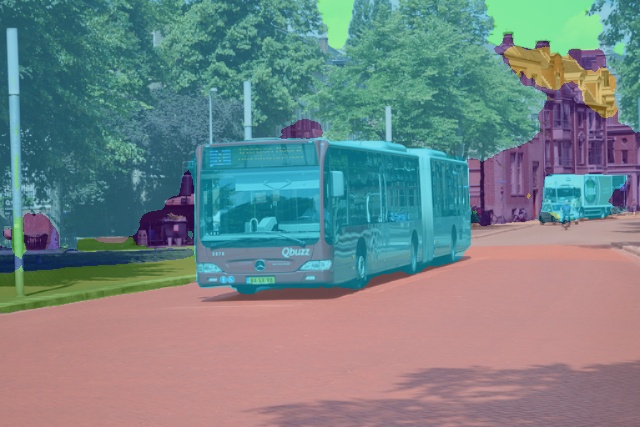}&
		\includegraphics[width=0.16\linewidth,height=0.8in,valign=m]{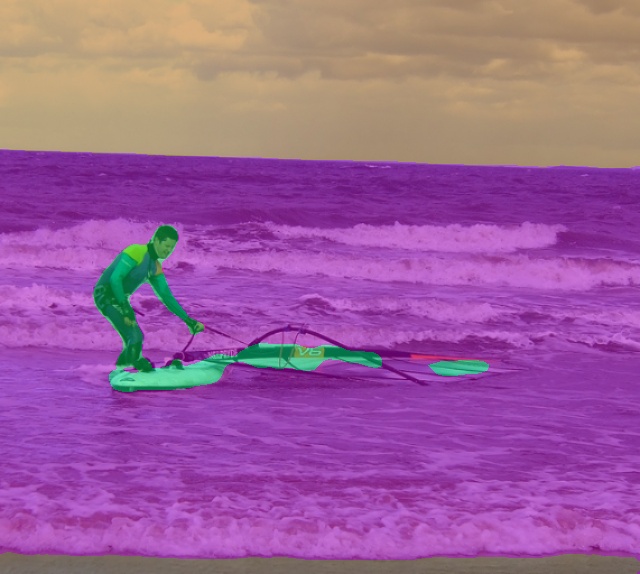} \\
	    \addlinespace[3pt]
	    \rotatebox[origin=c]{90}{\scriptsize{(d) Region Rebalance}}
        \includegraphics[width=0.16\linewidth,height=0.8in,valign=m]{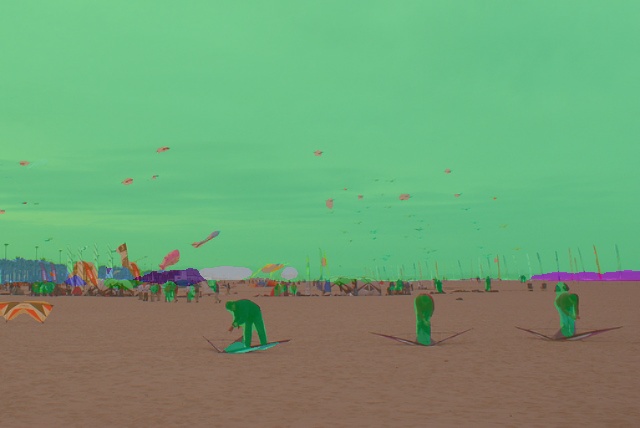}&
	    \includegraphics[width=0.16\linewidth,height=0.8in,valign=m]{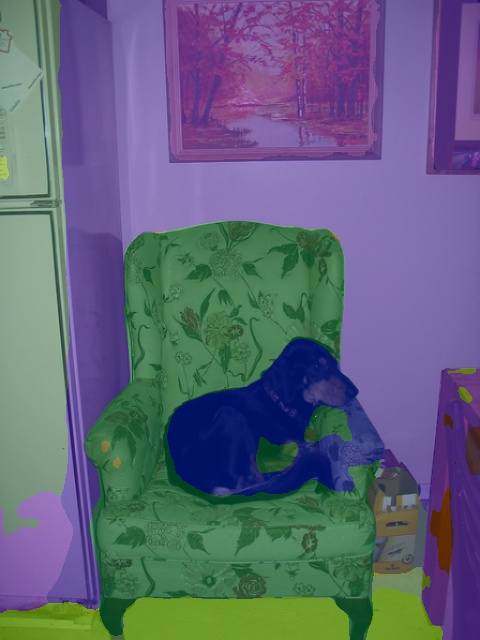}&
		\includegraphics[width=0.16\linewidth,height=0.8in,valign=m]{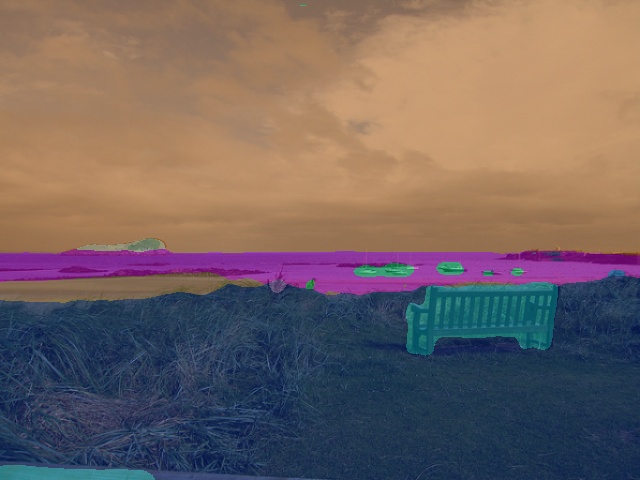}&
		\includegraphics[width=0.16\linewidth,height=0.8in,valign=m]{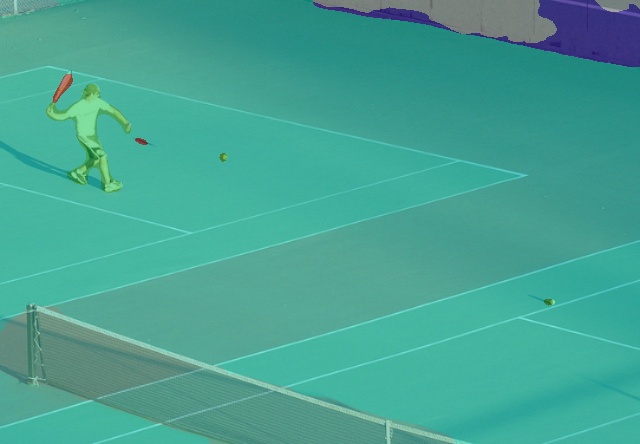}&
		\includegraphics[width=0.16\linewidth,height=0.8in,valign=m]{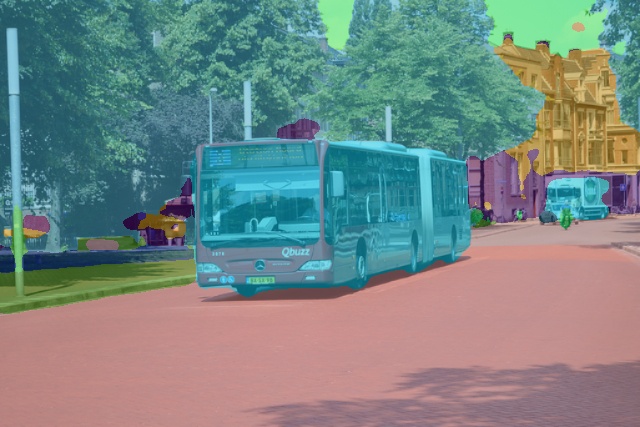}&
		\includegraphics[width=0.16\linewidth,height=0.8in,valign=m]{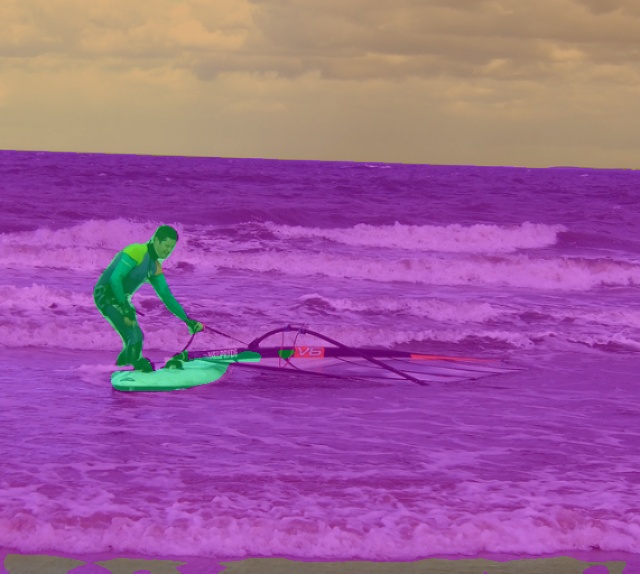} \\
	\end{tabular}}
	\caption{Visualization comparisons between baseline and region rebalance on COCO-Stuff$164$K.}
	\label{fig:visual_comparison_cocostuff_v1}
\end{figure*}

In this section, we demonstrate the advantages of region rebalance with quantitative visualizations on ADE$20$K and COCOStuff$164$K shwon in Figures \ref{fig:visual_comparison_ade20k_v1}, \ref{fig:visual_comparison_ade20k_v2} and \ref{fig:visual_comparison_cocostuff_v1}.

\end{document}